%% file: main.tex
\documentclass[preprint,12pt]{elsarticle}

\usepackage{amsmath,amssymb,amsfonts}
\usepackage{multirow}
\usepackage{array}
\usepackage{graphicx}
\usepackage{adjustbox}
\usepackage{xcolor}
\usepackage{tikz}
\usepackage{subcaption}
\usepackage{booktabs}
\usepackage{microtype}
\usepackage{algorithm}
\usepackage{xspace}
\usepackage{algpseudocode}
\usepackage{float}
\usepackage[most]{tcolorbox}
\usepackage[percent]{overpic}

\usepackage[breaklinks,colorlinks]{hyperref}

\graphicspath{{figures/}}

\usetikzlibrary{positioning,arrows.meta,calc,fit,backgrounds,matrix}
\definecolor{teal}{RGB}{0,128,128}

\usepackage{fontawesome5}
\newcommand{\ProjectShortName}{PG-3DGS\xspace}
\newcommand{\ProjectLongName}{\ProjectShortName: Optimizing 3D Gaussian Splatting to Satisfy Physics Objectives}
\newcommand{\ProjectMethodName}{Physics-Guided 3D Gaussian Splatting\xspace}

\definecolor{lightgray}{gray}{0.9}

\begin{document}

% \includepdf[fitpaper=true]{horizontal_graph_abs.pdf}

\begin{frontmatter}

\title{\ProjectLongName}

\author[inst1]{Zachary Lee\corref{cor1}}
\ead{lee5230@purdue.edu}
\author[inst1]{Maxwell Jacobson}
\ead{jacobs57@purdue.edu}
\author[inst1]{Yexiang Xue}
\ead{yexiang@purdue.edu}

\cortext[cor1]{Corresponding author}

% \address[inst1]{Department of Computer Science, Purdue University, West Lafayette, IN 47907, USA}
\affiliation[inst1]{organization={Department of Computer Science, Purdue University},
city={West Lafayette},
postcode={47907},
state={IN},
country={United States}}

\begin{abstract}
Recent advances in Gaussian Splatting have enabled fast, high-fidelity 3D scene generation, yet these methods remain purely visual and lack an understanding of how shapes behave in the physical world. We introduce \ProjectMethodName (\ProjectShortName), a framework that couples differentiable physics simulation with 3D Gaussian representations to generate 3D structures satisfying physics functionalities. By allowing physical objectives to guide the shape optimization process alongside visual losses, our approach produces geometries that are not only photometrically accurate but also physically functional. The model learns to adjust shapes so that the generated objects exhibit physically meaningful behaviors, for example, teapots that can pour and airplanes that can generate lift, without sacrificing visual quality. Experiments on pouring and aerodynamic lift tasks show that \ProjectShortName improves physical functionality while preserving visual quality. In addition to simulation gains, bench-top physical lift tests with 3D-printed aircraft (Cessna, B-2 Spirit, and paper plane) under identical airflow conditions show higher scale-measured lift for \ProjectShortName generated structures than an appearance-matching baseline in all three cases. Our unified framework connects appearance-based reconstruction with physics-based reasoning, enabling end-to-end generation of 3D structures that both look realistic and function correctly.
\end{abstract}

% \begin{graphicalabstract}
%     \includegraphics[angle=90, height=\textheight]{horizontal_graph_abs.pdf}
% \end{graphicalabstract}


\end{frontmatter}

% ================================================================
% 1. INTRODUCTION  (AIJ: Introduction)
% ================================================================
\section{Introduction}
\label{sec:intro}

\begin{figure}[H]
\centering

\includegraphics[width=\textwidth]{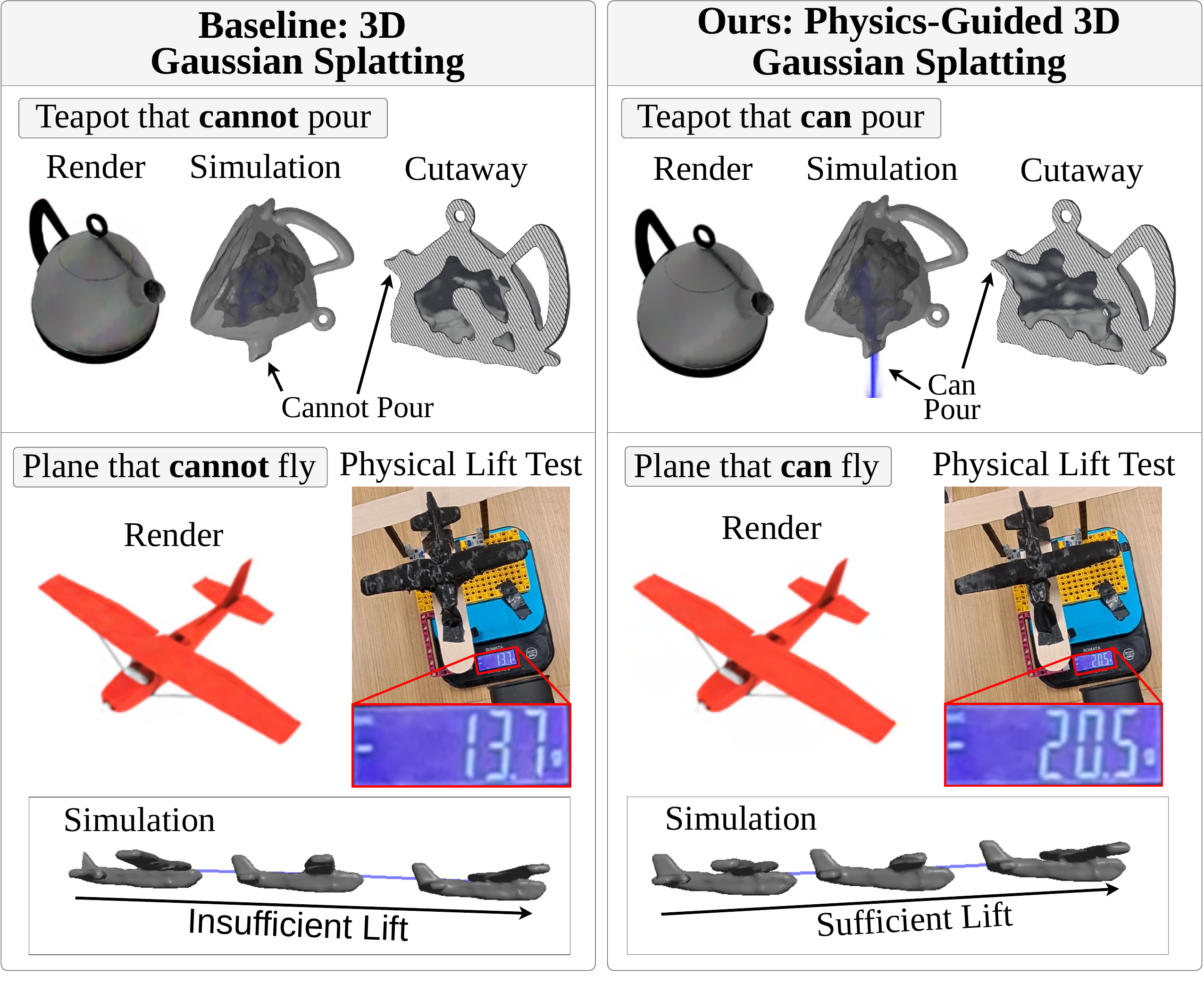}

\caption{Our proposed \ProjectMethodName (\ProjectShortName) is able to generate 3D structures that preserve appearance while inducing functional geometry.
Top: Prior work 3DGS produces a visually plausible teapot but lacks a functional cavity/spout connection, preventing pouring; our work \ProjectShortName recovers a connected cavity and opening, enabling successful pouring.
Bottom: Prior work 3DGS reconstruction yields a plane that cannot produce lift, whereas our work \ProjectShortName produces stable ascending flight.
}
\label{fig:teaser}
\end{figure}

Generative modeling has revolutionized AI, bringing us tools that generate vivid pictures \cite{yang2023diffusionsurvey}, detailed videos \cite{xing2024videosurvey}, and lively text \cite{LLMSurvey}. As AI is progressing towards modeling and interacting with our physical world \cite{ha2018worldmodels,ding2025understanding}, generative models have been migrating from generating 2D images to 3D structures. 
Reconstructing 3D objects from 2D images has become a practical primitive for computer vision, robotics, and embodied AI.
Recent neural and explicit radiance-field methods can infer detailed geometry and appearance from multi-view supervision, enabling
high-fidelity novel-view rendering and fast optimization of 3D representations
(e.g., NeRF~\cite{mildenhall_nerf_2020} and 3D Gaussian Splatting~\cite{kerbl_3d_2023}).
In parallel, text- and image-conditioned 3D generation has progressed rapidly by optimizing 3D representations through differentiable rendering
(e.g., DreamFusion/Magic3D-style score distillation) \cite{poole_dreamfusion_2022,lin_magic3d_2022}. 
A shared theme across these directions is \emph{inverse graphics}: when rendering is differentiable, gradients from image-space objectives can be back-propagated
to adjust a 3D representation so that its projections match observed images.

However, visually accurate 3D generation does not imply physical functionality.
Appearance-based optimization can converge to geometries that render correctly while violating basic physical requirements, such as
aerodynamics or internal structure.
Figure~\ref{fig:teaser} illustrates two representative failure modes.
In the airplane example, an appearance-only reconstruction matches the silhouette and texture yet produces wings that do not generate enough lift under a downstream physical test.
In the teapot example, an appearance-only reconstruction can look correct externally while producing a sealed interior or disconnected spout,
preventing fluid from pouring despite photometric agreement across views.
These failures matter whenever reconstructed objects are used beyond rendering: in simulation-based planning,
interactive scene understanding, or data-driven design of usable artifacts -- a system that cannot relate appearance and behavior cannot fulfill physical functionalities or enable realistic interactions in a virtual reality world.

This paper studies how to generate 3D structures that are functional in the physical world from 2D observations by optimizing physics objectives in addition to visual appearance matching. 
We introduce \textbf{\ProjectMethodName (\ProjectShortName)}, a framework that augments 3D Gaussian Splatting with differentiable physics optimization.
We cast functional 3D structure generation as joint optimization: the 3D reconstruction must closely align with the input images when projected into 2D, and at the same time, satisfy specified physical objectives.

\ProjectShortName~defines new ways to find the gradients from both visual and physics simulations. Then the geometry can be adjusted following gradients from both sources. 

To enable functional 3D structure generation, we propose a differentiable coupling between an explicit Gaussian representation and grid-based physics. 
Gaussian Splatting parameterizes a scene by a set of anisotropic Gaussian primitives whose parameters can be optimized efficiently~\cite{kerbl_3d_2023}.
Optimizing physics objectives in \ProjectShortName~requires us to tune these Gaussian parameters for physics objectives. 
To accomplish this, we construct an efficient and scalable \emph{soft occupancy / solid mask} on the simulation grid as a differentiable function of the Gaussian parameters.
This mediating field serves two purposes: (i) it converts the Gaussian geometry into grid-aligned coefficients that determine where Eulerian operators are applied and how fluid–solid boundary conditions are enforced, and (ii) gradients from physical objectives
(e.g., pressure, flow, net force/torque) can be propagated back through the mask to update Gaussian geometry.
We instantiate this idea with differentiable physical pipelines for incompressible fluid--solid interaction and rigid-body state evolution. In this case, 
3D generation tasks such as ``teapot successfully pouring'' or ``sustained flight'' can be learned in an end-to-end fashion. %from images.

Empirically, \ProjectShortName\ produces 3D reconstructions that remain visually competitive with appearance-only baselines while exhibiting substantially improved physical behavior.
On teapot instances, \ProjectShortName\ recovers connected cavities and openings that enable successful pouring.
On airplane instances, \ProjectShortName\ modifies wing geometry to increase lift, yielding stable flight trajectories under the same downstream physical test
while maintaining comparable novel-view rendering quality.
(Quantitative results and ablations appear in Sec.~\ref{sec:results}.) Beyond simulation-based evaluation, we also perform bench-top physical lift tests on 3D-printed aircraft under controlled airflow and observe consistently higher scale-measured lift for \ProjectShortName than an appearance-only baseline across three distinct shapes.

Beyond these specific tasks, we view \ProjectShortName\ as a reusable mechanism for \emph{physics-objective-driven 3D generation}. 
We anticipate that similar optimization loops can be applied to new domains (e.g., different PDEs, contact dynamics, or other functional criteria), 
given explicit differentiable 3D representations, differentiable mappings to simulator states, and a differentiable physical objective. 
We hope that our endeavor makes it easy for future work to swap in new physical objectives and simulators while reusing the same core coupling between 3D scene generation and physics simulation.

\section{Related Work}
\label{sec:related}
\subsection{3D Generative Models}
Recent advances in 3D neural generation have demonstrated that 
3D Gaussian Splatting (3DGS)~\cite{kerbl_3d_2023} achieves high-fidelity real-time radiance field reconstruction of complex scenes. Gaussian Splatting optimizes a set of anisotropic Gaussian primitives to approximate the 3D scenes.
Extensions such as 4D Gaussian Splatting~\cite{wu_4d_2023} capture dynamic scenes, 
DreamGaussian~\cite{tang_dreamgaussian_2023} introduces a generative model for efficient 3D content creation, 
and GS-IR~\cite{liang_gs-ir_2023} adapts Gaussian Splatting for inverse rendering and relighting.
Compared to implicit neural radiance fields (NeRF)~\cite{mildenhall_nerf_2020}, 
Gaussian-based representations are explicit, differentiable, and computationally efficient, 
allowing direct optimization of spatial position, opacity, and covariance.  
However, all of these methods focus exclusively on appearance-based optimization across multiple 2D views, 
without regard to the physical realism or functional behavior of the reconstructed geometry. More recently, there has been growing interest in \emph{physics-grounded} 3D assets that include physical properties beyond geometry and texture; for example, PhysX-3D introduces a large-scale dataset and generation paradigm with annotations such as absolute scale, material, affordances, and kinematics \cite{cao2025physx}.

\subsection{Differentiable Physics}
Differentiable physics frameworks enable gradients to flow through partial differential equation (PDE) solvers, 
allowing optimization of simulation parameters and control policies.
Particle-based methods such as SPNets~\cite{schenck_spnets_2018} and differentiable material point methods~\cite{xu_differentiable_2024} integrate fluid or deformable-body dynamics directly within deep networks.
Grid-based methods, including DiffTaichi~\cite{hu_difftaichi_2019} and 
the $\Phi$Flow framework~\cite{holl_phiflow_2020}, provide differentiable operators for advection, pressure projection, and diffusion.
Recent work has extended differentiable simulation to fluid--structure coupling, e.g.,
Differentiable Fluids with Solid Coupling~\cite{takahashi_differentiable_2021} 
and the differentiable pressure-implicit solver PICT~\cite{franz_pict_2025}.
These methods enable learning and inverse design for fluids, but the geometry is typically represented by static meshes or voxel grids.
Our formulation needs to incorporate differentiable physics simulation into 3D generative models such as 
Gaussian splatting. One challenge is in backpropagating the gradients from voxel or mesh-based  geometry into continuous Gaussian parameterized boundaries of solid–fluid interactions. We will discuss our approach to attacking this challenge in later sections. 

Physics-informed neural networks (PINNs) incorporate known physical structure into learning by representing the unknown solution field $u(x,t)$ with a neural network and penalizing violations of the governing PDE at collocation points, alongside any available supervision such as initial and boundary conditions \cite{raissi2019physics}. 
More broadly, PINNs can be viewed as a form of physics-informed machine learning in which physical laws act as informative priors and are enforced as soft penalty constraints through the training objective, improving data efficiency and discouraging physically implausible predictions \cite{karniadakis2021physics}. 
However, enforcing physics through composite losses can lead to challenging optimization behavior: stiff gradient-flow dynamics may induce severe imbalance between the PDE residual and data/constraint terms, resulting in unstable training and inaccurate solutions unless the loss is carefully balanced or adaptively reweighted \cite{wang2020understanding}. 
In our setting, we adopt the same core principle of using differentiable physics as a training signal, but apply it to guide optimization over our representation rather than directly regressing the full PDE solution field, leveraging physical consistency while avoiding some of the failure modes associated with residual-dominated training.

\subsection{Coupling Scene Representations with Physics}
Only a few works attempt to bridge neural scene representations and physical simulation.
The authors of~\cite{feng_gaussian_2024} unify 3D Gaussian Splatting with Position-Based Dynamics (PBD) to model fluid and solid motion within reconstructed 3DGS scenes,
achieving coherent forward simulations and view-consistent rendering.
However, it performs only forward dynamics and is not differentiable with respect to the Gaussian geometry parameters.
Our approach introduces a differentiable physics simulator integrated into the generative process,
enabling gradients to propagate directly to the underlying representation for 3D structure generation with physical functionality.
Recent work has begun integrating physics-based simulation with explicit scene representations for interactive dynamics.
PhysDreamer \cite{zhang2024physdreamer} represents objects with 3D Gaussians and optimizes a spatially-varying material field through differentiable MPM, leveraging priors distilled from video generation models to synthesize plausible object responses under novel interactions. Complementarily, PhysGen3D \cite{chen2025physgen3d} reconstructs an interactive 3D world from a single image and simulates future dynamics via physics-based simulation, enabling controllable physically grounded video generation.
These approaches primarily focus on inferring dynamics or material properties for interaction, whereas our method targets \emph{inverse design of geometry} by coupling a differentiable fluid solver with a Gaussian-induced boundary representation.

\subsection{Guidance-Based Optimization and Joint Objectives}
Our dual-objective optimization connects conceptually to the guidance mechanisms in diffusion models.  
Classifier guidance~\cite{dhariwal_diffusion_2021} and classifier-free guidance~\cite{ho_classifier-free_2022-1}
combine gradients from a base generative model and an auxiliary condition to trade off fidelity and diversity during sampling.
This principle descends from early score-based and diffusion formulations~\cite{sohl-dickstein_deep_2015,song_generative_2020,croitoru_diffusion_2023},
where auxiliary gradient terms steer a base generative process toward conditional targets.
Analogously, in our framework the visual reconstruction loss defines the base gradient (the image likelihood term), 
while the physics-based loss acts as a guidance gradient that steers geometry toward physically plausible and task-consistent configurations. This perspective is also reflected in recent physics-driven text-to-3D optimization methods, which refine diffusion-based shape priors using differentiable physics losses to enforce mechanical feasibility \cite{xu2024precise}. This probabilistic view naturally connects our method to a Bayesian posterior over 3D shapes conditioned on both visual and physical evidence (Sec.~\ref{sec:problem}).

\section{Preliminaries}
\label{sec:prelim}

\subsection{Notation and setting}
\label{subsec:prelim_notation}

For all simulations, we consider a bounded three-dimensional domain $\Omega \subset \mathbb{R}^3$ discretized as a regular voxel grid
$\Omega_h$ with spacing $h$.
Time is discretized into steps $t \in \{0,\dots,T\}$ with timestep $\Delta t$.

In this work, our goal is to create 3D objects that are both visually coherent and meet physics 
specifications -- such as teapots that can hold and pour water, and airplanes that can produce lift. We focus on specifications over fluid dynamics in this work, but the approach is generalizable to many physics frameworks.
Fluid state variables at time $t$ consist of a velocity field
$\mathbf{u}_t : \Omega_h \rightarrow \mathbb{R}^3$ and a pressure field
$p_t : \Omega_h \rightarrow \mathbb{R}$. Each $\mathbf{u}_t$ is a tuple $\mathbf{u}_t = (u_{tx}, u_{ty}, u_{tz})$, representing the velocity of the fluid along the three dimensions, respectively. 
For simulations involving transported quantities (e.g., a scalar smoke or dye density used to track fluid motion), we additionally maintain a scalar field
$\rho_t : \Omega_h \rightarrow \mathbb{R}$.

The pressure field is a scalar field that enforces the incompressibility of the fluid. Spatial variations in pressure generate accelerations that redistribute momentum and prevent local compression or expansion of the fluid. In this sense, pressure acts as a constraint force internal to the fluid, shaping the velocity field rather than evolving independently.

To model interactions between a rigid solid and the surrounding flow—such as an aircraft accelerating during takeoff—we maintain a rigid-body state describing the object’s pose and motion within the fluid. At each time step \(t\), this state consists of a position \(\mathbf{x}_t \in \mathbb{R}^3\) giving the center-of-mass location in world coordinates, a linear velocity \(\mathbf{v}_t \in \mathbb{R}^3\) specifying translational motion along the three spatial axes, an orientation \(\mathbf{q}_t \in \mathbb{H}\) represented as a unit quaternion, and an angular momentum \(\boldsymbol{L}_t \in \mathbb{R}^3\).

The orientation quaternion \(\mathbf{q}_t = (q_w, q_x, q_y, q_z)\) consists of a scalar component \(q_w \in \mathbb{R}\) and a vector component \((q_x, q_y, q_z) \in \mathbb{R}^3\). When constrained to unit norm, this representation encodes a three-dimensional rotation by an angle \(\varphi\) about a unit axis \(\mathbf{n}\), via
\[
q_w = \cos(\varphi/2), \qquad (q_x, q_y, q_z) = \mathbf{n}\sin(\varphi/2).
\]
Unit quaternions are a standard representation for tracking orientation in rigid-body dynamics and computer graphics, as they avoid the singularities of Euler angles, enable stable composition of rotations through quaternion multiplication, and support smooth, differentiable time integration of rotational motion.

Given the current orientation \(\mathbf{q}_t\), the rigid body has world-frame inertia
\[
\mathbf I_t^{w} = R(\mathbf q_t)\,\mathbf I_{\mathrm{body}}\,R(\mathbf q_t)^\top,
\]
where \(\mathbf I_{\mathrm{body}} \in \mathbb{R}^{3\times 3}\) is the constant inertia tensor in the
body frame, i.e., the coordinate frame attached to the object and aligned with it as it moves and
rotates. Rather than storing angular velocity directly, we represent rotational motion using the
world-frame angular momentum \(\mathbf L_t \in \mathbb{R}^3\), and recover angular velocity as
\[
\boldsymbol\omega_t = (\mathbf I_t^{w})^{-1}\mathbf L_t.
\]

Together, \((\mathbf{x}_t, \mathbf{q}_t)\) define the rigid-body pose, while \(\mathbf{v}_t\) and
\(\mathbf L_t\) define its translational and rotational motion state. These variables are defined
for all simulations; in experiments involving static objects, rigid-body motion is disabled by
fixing \(\mathbf{v}_t = \mathbf{0}\) and \(\mathbf L_t = \mathbf{0}\) for all \(t\).

Visual observations are given by a set of calibrated images
\(\mathcal{I} = \{ I_k \}_{k=1}^K\) with known camera models
\(\{ \pi_k \}_{k=1}^K\). Each camera model \(\pi_k\) specifies a deterministic
mapping from three-dimensional world coordinates to two-dimensional image
coordinates. This mapping is defined by camera extrinsic parameters, which
describe the rigid transformation from the world frame to the camera frame,
and intrinsic parameters, which describe the perspective projection from
camera coordinates to pixel coordinates. The extrinsics consist of a rotation
\(\mathbf{R}_k \in \mathbb{R}^{3\times 3}\) and translation
\(\mathbf{t}_k \in \mathbb{R}^3\), while the intrinsics include focal lengths and
a principal point that together define the camera calibration matrix. The
composition of these components yields the full projection \(\pi_k\).
All learnable geometric parameters are collectively denoted by \(\theta\). In
this work, \(\theta\) corresponds to the parameters of the 3D Gaussian scene
representation, including Gaussian means, covariances, opacities, and
appearance coefficients, which are defined formally in
Sec.~\ref{subsec:prelim_gaussians}.

\subsection{3D Gaussian scene representation}
\label{subsec:prelim_gaussians}

We represent scene geometry using a collection of $N$ anisotropic 3D Gaussians,
following the explicit representation introduced in 3D Gaussian Splatting~\cite{kerbl_3d_2023}.
Each Gaussian $i$ is parameterized by
\[
(\boldsymbol{\mu}_i, \Sigma_i, \alpha_i, \mathbf{c}_i),
\]
where $\boldsymbol{\mu}_i \in \mathbb{R}^3$ denotes the Gaussian mean (center position),
$\Sigma_i \in \mathbb{R}^{3 \times 3}$ is a positive-definite covariance matrix encoding
its spatial extent and orientation, $\alpha_i \in [0,1]$ is an opacity value controlling
its contribution to rendering, and $\mathbf{c}_i$ are view-dependent appearance coefficients, which determine the color of a Gaussian based on a viewing angle. In essence, the scene or object to model is represented by a collection of many ellipsoids that each have their own position, size, orientation, and color, and whose opacity tapers off according to a Gaussian PDF (multiplied by a given base opacity).

\paragraph{Covariance parameterization}
To ensure $\Sigma_i$ remains symmetric and positive definite during optimization,
we parameterize it using a rotation--scale decomposition:
\begin{equation}
\Sigma_i = R(\mathbf{q}_i)\,\mathrm{diag}(\mathbf{s}_i^2)\,R(\mathbf{q}_i)^\top,
\end{equation}
where $\mathbf{q}_i$ is a unit quaternion representing the Gaussian's orientation and
$\mathbf{s}_i \in \mathbb{R}_+^3$ are axis-aligned scale parameters.
Intuitively, this corresponds to starting from an axis-aligned ellipsoid with principal
radii $\mathbf{s}_i$, and rotating it in space according to $\mathbf{q}_i$.
This parameterization avoids invalid covariances while allowing anisotropic,
oriented Gaussian primitives to adapt to local scene geometry.

\paragraph{Spherical harmonics (view-dependent appearance)}
\label{par:prelim_sh}

In a 3D Gaussian scene representation, each Gaussian must specify what it looks like when viewed by a camera.
A constant RGB color per Gaussian is often too restrictive: the same surface point can appear
different across viewpoints due to effects such as specular highlights, Fresnel-like reflectance,
and other non-Lambertian behavior.
To capture this without introducing an expensive material model, 3D Gaussian Splatting represents
color as a simple smooth function of viewing direction.

Let $\mathbf d\in\mathbb S^2$ be the unit viewing direction (from the Gaussian toward the camera).
We parameterize the RGB appearance of Gaussian $i$ using a truncated spherical harmonics (SH)
expansion:
\begin{equation}
\mathbf c_i(\mathbf d)
=
\sum_{\ell=0}^{L}\sum_{m=-\ell}^{\ell}
\mathbf a^{(i)}_{\ell m}\,Y_{\ell m}(\mathbf d),
\label{eq:sh_color}
\end{equation}
where $Y_{\ell m}(\mathbf d)$ are fixed basis functions on the unit sphere and
$\mathbf a^{(i)}_{\ell m}\in\mathbb R^3$ are learnable coefficients.
Using a low maximum degree $L$ provides a compact representation with $(L+1)^2$ coefficients per
color channel: $\ell=0$ corresponds to a view-independent base color, while higher-order terms allow
gradual view-dependent changes.

In summary, spherical harmonics provide a lightweight way to model view-dependent appearance in a
continuous and differentiable form, improving multi-view consistency while keeping the scene
representation compact.

\subsection{Eulerian incompressible fluid simulation}
\label{subsec:prelim_fluids}

Fluid motion is modeled using the incompressible Navier--Stokes equations,
\begin{align}
\underbrace{\frac{\partial \mathbf{u}}{\partial t}}_{\text{temporal change}}
\;+\;
\underbrace{(\mathbf{u}\cdot\nabla)\mathbf{u}}_{\text{advection}}
&=
\underbrace{-\nabla p}_{\text{pressure}}
\;+\;
\underbrace{\mathbf{f}}_{\text{external forces}}, \label{eq:ns_momentum} \\
\nabla \cdot \mathbf{u} &= 0, \qquad \text{(incompressibility)} \label{eq:ns_incompressibility}
\end{align}

where $\mathbf{u}(\mathbf{x},t)$ is the velocity field, $p(\mathbf{x},t)$ is
pressure, and $\mathbf{f}(\mathbf{x},t)$ denotes external body forces.
These equations describe conservation of momentum for an incompressible fluid
and are standard in computer graphics and physics-based simulation
\cite{bridson_2016}.

\paragraph{Material derivative and advection}
The nonlinear term $(\mathbf{u}\cdot\nabla)\mathbf{u}$ arises from the
\emph{material derivative}, which measures the rate of change of velocity
as fluid particles move through space. In an Eulerian formulation, this effect
is handled explicitly via \emph{advection}: velocities are transported along
the flow field itself. Advection is therefore the mechanism by which motion is
carried forward in time and is essential for capturing convection and transport
phenomena in fluid dynamics.

\paragraph{Discrete projection method}
We discretize the fluid domain using a regular Eulerian grid $\Omega_h$ and
advance the velocity field using a standard projection method
\cite{chorin_projection_1968, stam_stable_1999}. Each timestep splits the update
into conceptually distinct stages corresponding to terms in
Eq.~\eqref{eq:ns_momentum}--\eqref{eq:ns_incompressibility}.

Given the velocity field $\mathbf{u}_t$ at time $t$, we first compute an
intermediate velocity $\tilde{\mathbf{u}}$ by advecting the velocity field and
applying external forces:
\begin{align}
\tilde{\mathbf{u}} = \mathcal{A}(\mathbf{u}_t) + \Delta t\,\mathbf{f},
\end{align}
where $\mathcal{A}$ denotes the advection operator.
We then enforce incompressibility by subtracting the gradient of a pressure
field:
\begin{align}
\mathbf{u}_{t+1} = \tilde{\mathbf{u}} - \nabla p,
\end{align}
where $p$ is obtained by solving a discrete Poisson equation derived from
$\nabla \cdot \mathbf{u}_{t+1} = 0$.

\paragraph{Advection operator}
Advection is implemented using semi-Lagrangian backtracing: for each grid cell,
we trace a particle backward in time along the velocity field and sample the
previous velocity using trilinear interpolation. This scheme is unconditionally
stable and widely used in graphics-oriented fluid solvers
\cite{stam_stable_1999}.

\section{Problem Formulation}
\label{sec:problem}

We consider the problem of reconstructing a three-dimensional object from visual observations
such that the recovered geometry is also compatible with desired physical behavior.
Let $\mathcal I=\{I_k\}_{k=1}^K$ denote a set of calibrated images of the object.

The unknown is a set of geometric parameters $\theta$ defining a 3D object.
In this work, $\theta$ will later be instantiated as the parameters of a 3D Gaussian
representation, though the formulation in this section is agnostic to the particular
geometric parameterization.

In addition to matching the visual observations, we require the reconstructed geometry to
perform well under a task-specific physical objective.
We represent this objective by a functional $\Phi$ acting on the physical trajectory induced
by the geometry. Denoting this trajectory by $\tau(\theta)$, the quantity
$\Phi(\tau(\theta))$ measures task-specific physical compatibility.
Depending on the application, $\Phi$ may evaluate criteria such as poured volume, lift,
weighted density, average vertical position, or final pose.

To combine visual consistency, physical behavior, and geometric regularization in a single
probabilistic formulation, we define a Gibbs posterior over geometry parameters by
\begin{equation}
p(\theta \mid \mathcal I,\Phi)
=
\frac{\exp\!\big(-\mathcal J(\theta;\mathcal I,\Phi)\big)}
{Z(\mathcal I,\Phi)},
\label{eq:gibbs_posterior}
\end{equation}
where
\begin{equation}
Z(\mathcal I,\Phi)
=
\int \exp\!\big(-\mathcal J(\theta';\mathcal I,\Phi)\big)\, d\theta'
\label{eq:gibbs_partition}
\end{equation}
is the normalizing constant.

The objective $\mathcal J(\theta;\mathcal I,\Phi)$ is decomposed as
\begin{equation}
\mathcal J(\theta;\mathcal I,\Phi)
=
\mathcal L_{\mathrm{vis}}(\theta;\mathcal I)
+
\mathcal L_{\mathrm{phys}}(\theta;\Phi)
+
\mathcal L_{\mathrm{reg}}(\theta),
\label{eq:problem_joint_objective}
\end{equation}
where $\mathcal L_{\mathrm{vis}}$ measures compatibility with the observed images,
$\mathcal L_{\mathrm{phys}}$ measures compatibility with the desired physical behavior, and
$\mathcal L_{\mathrm{reg}}(\theta)$ encodes prior preferences or regularization on the geometry.
In particular, the physical term is defined abstractly through the task functional as
\begin{equation}
\mathcal L_{\mathrm{phys}}(\theta;\Phi)=\Phi(\tau(\theta)).
\label{eq:abstract_phys}
\end{equation}

Our target estimate is the maximum a posteriori solution
\begin{equation}
\theta^\star
=
\arg\max_\theta p(\theta\mid \mathcal I,\Phi).
\label{eq:map}
\end{equation}
Since $Z(\mathcal I,\Phi)$ does not depend on $\theta$, MAP inference is equivalent to
minimizing the objective
\begin{equation}
\theta^\star
=
\arg\min_\theta \mathcal J(\theta;\mathcal I,\Phi).
\label{eq:map_equiv}
\end{equation}

The remainder of the paper specifies how the geometry is represented, how the visual and
physical terms are instantiated, and how the resulting objective is optimized in a
differentiable manner.

\section{\ProjectMethodName}
\label{sec:method}
\subsection{Overview of the optimization pipeline}
\label{subsec:method_overview}

\begin{algorithm}[t]
\caption{Joint optimization with simulation-driven physical objectives}
\label{alg:joint_opt}
\begin{algorithmic}[1]
\Require Images $\{I_k\}_{k=1}^K$, cameras $\{\pi_k\}_{k=1}^K$, initial Gaussians $\theta_0$, Physical Objective $\Phi$
\For{$n = 0,\dots,N_{\mathrm{iter}}-1$}
    \State Render $\hat I_k \gets \mathcal R(\theta_n;\pi_k)$ for $k=1,\dots,K$
    \State Compute visual loss $\mathcal L_{\mathrm{vis}}$ from $\{\hat I_k\}_{k=1}^K$ and $\mathcal I$
    \algblock{Phys}{EndPhys}
    \algnewcommand\algorithmicphys{\textbf{Compute $\mathcal L_{\mathrm{phys}}$:}}
    \algnewcommand\algorithmicendphys{\textbf{end}}
    \algrenewtext{Phys}{\algorithmicphys}
    \algrenewtext{EndPhys}{\algorithmicendphys}
    \Phys
    \State Convert Gaussians to mask for simulator $\chi(\cdot;\theta_n)$ (Sec.~\ref{subsec:method_mask})
    \State Differentiably simulate and obtain results $\tau$
    (Secs.~\ref{subsec:method_sim}--\ref{subsec:method_rigidbody})
    \State Compute $\mathcal L_{\mathrm{phys}}=\Phi(\tau)$ from objective
    (Sec.~\ref{subsec:method_sim})
    \EndPhys
    \State Form total loss $\mathcal J(\theta_n)$ (Sec.~\ref{subsec:method_jointopt})
    \State Update $\theta_{n+1} \gets \theta_n - \eta \nabla_\theta \mathcal J(\theta_n)$
\EndFor
\State \Return optimized Gaussians $\theta^\star = \theta_{N_{\mathrm{iter}}}$
\end{algorithmic}
\end{algorithm}

Given a set of calibrated multi-view images $\{I_k\}_{k=1}^K$ with camera parameters
$\{\pi_k\}_{k=1}^K$, we optimize a 3D Gaussian scene representation parameterized by $\theta$
(Sec.~\ref{subsec:prelim_gaussians}).
Optimization proceeds by minimizing the joint objective $\mathcal J(\theta)$ from Equation~\ref{eq:problem_joint_objective} using gradient-based updates.

At each iteration, we evaluate two complementary supervision signals.
First, a visual loss $\mathcal L_{\mathrm{vis}}$ is computed by differentiably rendering the current
Gaussian representation into predicted images
$\hat I_k=\mathcal R(\theta;\pi_k)$ and comparing them to the input views.
Second, a physical loss $\mathcal L_{\mathrm{phys}}$ is computed by running a differentiable
fluid--solid simulation in which the solid boundary is induced by $\theta$ through a
Gaussian-to-mask mapping.

These two losses, together with optional regularization, are combined into the total objective
$\mathcal J(\theta)$, whose gradient is backpropagated end-to-end to update the Gaussian parameters.
Algorithm~\ref{alg:joint_opt} summarizes this optimization procedure.

\begin{figure}[H]
\centering
% \includestandalone[width=\linewidth]{method_flowchart}
\input{method_flowchart_raw}
\caption{Method overview: Visual supervision uses calibrated images and cameras; physics supervision uses a simulation and task specification encoding desired physical behavior. Both contribute gradients w.r.t.\ the same Gaussian parameters $\theta$.}
\label{fig:method_overview}
\end{figure}
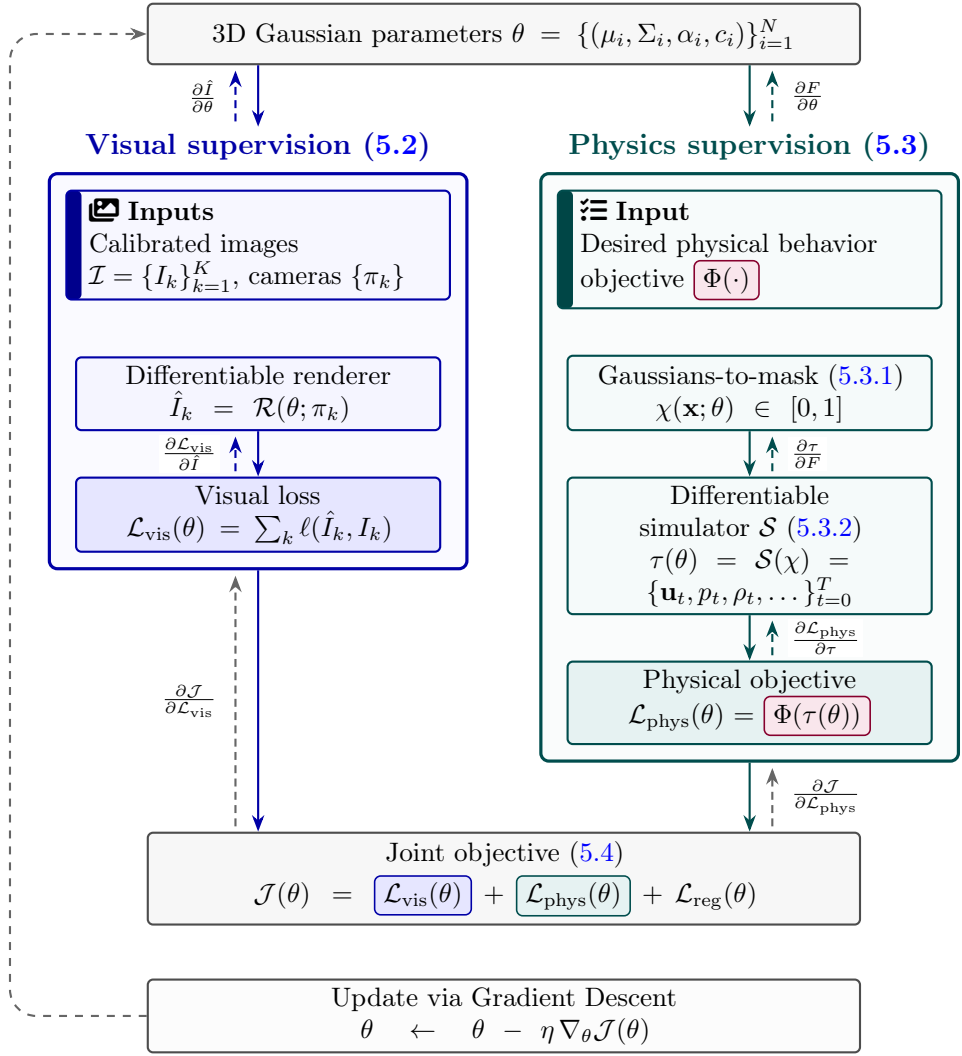

Figure~\ref{fig:method_overview} provides a high-level overview of the data flow.
The key technical components introduced in the following sections are:
(1) a differentiable mapping from continuous Gaussian primitives to a grid-aligned solid mask
(Sec.~\ref{subsec:method_mask}), and
(2) differentiable fluid--solid coupling based on this soft mask that preserves gradients through
boundary interactions (Sec.~\ref{subsec:method_brinkman}).

\subsection{Visual Supervision}
\label{subsec:visual_supervision}

Visual supervision encourages the Gaussian representation $\theta$, once rendered, to match the observed multi-view
images $\mathcal I=\{I_k\}_{k=1}^K$. The key insight \cite{mildenhall_nerf_2020} and \cite{kerbl_3d_2023} made was that the rendering process itself can be a differentiable function of our scene representation. This allows us to use standard gradient descent methods to optimize our Gaussian parameters in order to minimize the difference between our rendered views of our Gaussians and ground truth images $\mathcal I$.
Given camera parameters $\pi_k$, we render predicted images
$\hat I_k = \mathcal R(\theta;\pi_k)$ using the 3D Gaussian Splatting renderer~\cite{kerbl_3d_2023},
and define the visual reconstruction loss
\begin{equation}
\mathcal{L}_{\mathrm{vis}}(\theta)
= \sum_{k=1}^K \ell\!\left(\hat I_k, I_k\right),
\label{eq:lvis}
\end{equation}
where $\ell(\cdot,\cdot)$ is a per-image discrepancy measure. For our purposes, a combination of $\ell_1$ loss and SSIM is used.

\paragraph{Differentiable rendering}
The renderer $\mathcal R(\theta;\pi_k)$ follows the 3D Gaussian Splatting formulation \cite{kerbl_3d_2023}, which represents
the scene as a set of anisotropic 3D Gaussian primitives but performs rendering entirely in image
space. Each Gaussian $i$ is parameterized by a 3D mean $\boldsymbol{\mu}_i$, covariance
$\Sigma_i \in \mathbb{R}^{3\times 3}$, color $\mathbf{c}_i$ (assuming a given viewing direction), and opacity $\alpha_i$.

Given camera parameters $\pi_k$, each 3D Gaussian is mapped to the image plane through the nonlinear
camera projection $\mathbf{u} = \Pi(\mathbf{x}; \pi_k)$. Projecting a 3D Gaussian to 2D through a pinhole camera does not have a simple form, so to obtain a tractable screen-space
representation, the projection is locally linearized around the Gaussian mean
$\boldsymbol{\mu}_i$. Specifically, letting
\[
\mathbf{J}_i = \left.\frac{\partial \Pi(\mathbf{x}; \pi_k)}{\partial \mathbf{x}}\right|_{\mathbf{x}=\boldsymbol{\mu}_i}
\]
denote the Jacobian of the projection at $\boldsymbol{\mu}_i$, the projected 2D mean and covariance
are given by
\[
\boldsymbol{\mu}_i^{(2D)} = \Pi(\boldsymbol{\mu}_i; \pi_k),
\qquad
\Sigma_i^{(2D)} = \mathbf{J}_i \,\Sigma_i\, \mathbf{J}_i^\top.
\]
This first-order approximation maps the anisotropic 3D Gaussian to a 2D Gaussian in screen space, whose extent and orientation reflect both the original 3D covariance and the local
geometry of the projection. This results in a
screen-space density
\[
G_i(\mathbf{u})
=
\exp\!\left(
-\tfrac{1}{2}
(\mathbf{u}-\boldsymbol{\mu}_i^{(2D)})^\top
(\Sigma_i^{(2D)})^{-1}
(\mathbf{u}-\boldsymbol{\mu}_i^{(2D)})
\right),
\]
which determines the spatial extent of the Gaussian’s contribution to nearby pixels.

Rendering proceeds by sorting Gaussians by depth and compositing their screen-space contributions
using front-to-back alpha blending. For a pixel location $\mathbf{u}$, the rendered color is
\[
\hat I_k(\mathbf{u})
=
\sum_i
T_i(\mathbf{u})\,
\alpha_i\,G_i(\mathbf{u})\,\mathbf{c}_i,
\qquad
T_i(\mathbf{u})
=
\prod_{j<i}\bigl(1-\alpha_j\,G_j(\mathbf{u})\bigr),
\]
where $T_i(\mathbf{u})$ denotes the accumulated transmittance at that pixel due to previously composited Gaussians, since we desire Gaussians in front to block Gaussians behind them to an extent.

All steps of this process—3D-to-2D projection, Gaussian covariance transformation, screen-space
evaluation, and alpha compositing—are differentiable with respect to the Gaussian parameters. As a
result, image reconstruction errors backpropagate directly to 3D positions, shapes, opacities, and
colors, enabling efficient end-to-end optimization from multi-view image supervision.

\subsection{Physics Supervision}
\label{subsec:method_sim}

Physics supervision encourages the reconstructed geometry to exhibit task-specific physical behavior
when embedded in a differentiable fluid--solid simulator.
We define the physical loss as a differentiable function of the simulated trajectory,
\begin{equation}
\mathcal L_{\mathrm{phys}}(\theta) = \Phi\big(\tau(\theta)\big),
\label{eq:phys_functional}
\end{equation}
where $\Phi(\cdot)$ encodes the task objective. As an example, a task objective for increasing lift may involve simply summing the $z$ positions of the solid that are output from the simulation, and negating the result to form a minimization problem. Gradients would then flow back from positions to forces, from forces to fluid interactions etc., and finally to geometry at each timestep to increase the average $z$ position after optimization.
We describe concrete instantiations of $\Phi$ in Sec.~\ref{sec:experiments}.

Evaluating $\mathcal L_{\mathrm{phys}}$ requires unrolling a differentiable simulator.
Let $\mathcal \tau$ denote the simulator operator; given geometry parameters $\theta$, it produces a
trajectory of fluid (and optionally rigid-body) states over a horizon $T$:
\begin{equation}
\tau(\theta)
= \mathcal S (\chi(\theta)) =
\Big\{
\mathbf u_t,\; p_t,\; \rho_t,\; \mathbf x_t,\; \mathbf v_t,\; \mathbf q_t,\; \boldsymbol L_t
\Big\}_{t=0}^{T}.
\label{eq:trajectory_def}
\end{equation}

Simulator state variables produced by the simulator are summarized in Table~\ref{tab:sim_states}. 

\begin{table}[h]
\centering
\begin{tabular}{{l p{0.7\linewidth}}}
\toprule
Symbol & Description \\
\midrule
$\mathbf u_t$ & Fluid velocity field at time $t$ ($\mathbf u_t : \Omega_h \rightarrow \mathbb{R}^3$) \\
$p_t$ & Fluid pressure field at time $t$ ($p_t : \Omega_h \rightarrow \mathbb{R}$) \\
$\rho_t$ & Transported scalar field (e.g., dye/smoke density used to track fluid motion) at time $t$ ($\rho_t : \Omega_h \rightarrow \mathbb{R}$) \\
$\mathbf x_t$ & Rigid-body center-of-mass position at time $t$ \\
$\mathbf v_t$ & Rigid-body linear velocity at time $t$ \\
$\mathbf q_t$ & Rigid-body orientation represented as a unit quaternion at time $t$ \\
$\boldsymbol L_t$ & Rigid-body angular momentum vector at time $t$ \\
\bottomrule
\end{tabular}
\caption{State variables produced by the differentiable simulator.}
\label{tab:sim_states}
\end{table}

A key challenge is that the simulator operates on grid-aligned fields over $\Omega_h$, while geometry
is represented as a continuous set of 3D Gaussians.
We therefore first construct a differentiable mapping from $\theta$ to a grid-based soft solid mask,
which serves as the interface through which geometry influences the discrete fluid operators.

\subsubsection{Gaussian-to-grid coupling via a soft solid mask}
\label{subsec:method_mask}

The fluid simulator operates on grid-aligned fields defined over the discretized domain $\Omega_h$,
whereas the scene geometry is represented as a continuous collection of anisotropic 3D Gaussians
(Sec.~\ref{subsec:prelim_gaussians}).
To couple these representations, we construct a differentiable mapping from Gaussian parameters
$\theta$ to a grid-based solid--fluid mask $\chi (\theta)$ that mediates fluid--solid interactions while preserving
gradient flow with respect to $\theta$.

\paragraph{Gaussian occupancy field}
Each Gaussian $i$ induces a spatial contribution at location $\mathbf x\in\Omega$ given by
\begin{equation}
\hat{\alpha}_i(\mathbf x)
=
\alpha_i
\exp\!\left(
-\tfrac{1}{2}(\mathbf x-\mu_i)^\top \Sigma_i^{-1}(\mathbf x-\mu_i)
\right),
\end{equation}
where $\mu_i$, $\Sigma_i$, and $\alpha_i$ denote the mean, covariance, and opacity of the Gaussian.
This quantity can be interpreted as a soft, continuous occupancy contribution and is differentiable
with respect to all Gaussian parameters.

\paragraph{Probabilistic union of Gaussians}
To aggregate individual contributions into a global solid representation, we adopt a probabilistic
union that maps multiple overlapping Gaussians to a single occupancy field:
\begin{equation}
\chi(\mathbf x;\theta)
=
1 - \prod_{i=1}^N \big(1 - \hat{\alpha}_i(\mathbf x)\big),
\label{eq:solid_occupancy}
\end{equation}
where $\chi(\mathbf x;\theta)\in[0,1]$ denotes the solid occupancy at location $\mathbf x$.
This construction smoothly interpolates between empty space ($\chi\approx 0$) and solid regions
($\chi\approx 1$), is permutation-invariant with respect to the Gaussian set, and remains fully
differentiable with respect to $\theta$. 
This mask is evaluated at grid cell centers $\mathbf x\in\Omega_h$ and serves as the interface
through which geometry influences the discrete fluid operators.

\paragraph{Efficient mask evaluation}
Na\"ively evaluating~\eqref{eq:solid_occupancy} at all grid cells scales as $O(N|\Omega_h|)$, which is
prohibitive for large $N$.
We therefore exploit the locality of Gaussian support and evaluate the mask using a tiled
rasterization-style procedure: each Gaussian is restricted to a conservative grid-aligned bounding
region, and each tile accumulates contributions only from Gaussians that intersect it.
This acceleration leaves the definition of $\chi(\mathbf x;\theta)$ unchanged; full details and
pseudocode are provided in \ref{app:mask_tiling}.

\paragraph{Differentiability}
Unlike binary voxelizations, which introduce discontinuities when geometry crosses grid boundaries,
the soft mask $\chi(\mathbf x;\theta)$ varies smoothly with the Gaussian parameters.
As a result, all downstream quantities that depend on the mask—such as solid–fluid coupling forces
introduced via Brinkman penalization—remain differentiable with respect to $\theta$.

This construction enables gradient-based inverse design through grid-based
fluid simulation.

\subsubsection{Fluid Simulation}
\label{subsec:method_fluid_sim}
\paragraph{Governing equations}
The simulator $\mathcal S(\chi)$ evolves an incompressible Newtonian fluid interacting with a rigid solid on a fixed Eulerian grid.
We model the interaction using a Brinkman penalization term (described further in Sec.~\ref{subsec:method_brinkman}) defined by a soft solid indicator
$\chi(\mathbf{x};\theta)\in[0,1]$ induced by the Gaussian geometry.
The resulting equations are
\begin{align}
\frac{\partial \mathbf{u}}{\partial t} + (\mathbf{u}\cdot\nabla)\mathbf{u}
&= -\nabla p + \mathbf{f}
+ \lambda\,\chi(\mathbf{x};\theta)\big(\mathbf{u}_{\mathrm{solid}}(\mathbf{x})-\mathbf{u}(\mathbf{x})\big),
\label{eq:method_ns_brinkman}\\
\nabla\cdot\mathbf{u} &= 0,
\label{eq:method_ns_incompressibility}
\end{align}
where $\mathbf{u}$ is velocity, $p$ is pressure, $\mathbf{f}$ includes external forces such as gravity, and $\lambda$ controls the strength of the Brinkman term.

\paragraph{Why soft boundaries are needed}
In many grid-based incompressible solvers, solid geometry is represented by a binary indicator
(e.g., a fluid mask $F:\Omega_h\!\to\!\{0,1\}$ with $F=1$ in fluid cells and $F=0$ in solid cells).
This binary discretization induces discontinuous changes in the set of active fluid cells as the
geometry moves across voxel boundaries, making the simulated outcome non-smooth as a function of
geometry parameters.

Immersed-boundary approaches such as Brinkman penalization similarly rely on a solid indicator to
apply a damping force inside the solid region. When this indicator is binary, the mapping from
geometry to simulator coefficients is still discontinuous, which yields vanishing or undefined
gradients almost everywhere with respect to geometry.

To enable gradient-based inverse design, we instead use the \emph{soft} mask
$\chi(\mathbf x;\theta)\in[0,1]$ described in Sec.~\ref{subsec:method_mask} that smoothly interpolates between fluid and solid regions.
Solid--fluid interaction is enforced via a differentiable Brinkman penalization term that depends
continuously on $\chi$.
In practice, the mask transition width controls a tradeoff: sharper masks more closely approximate
an impermeable boundary, while smoother masks provide better-conditioned gradients by distributing
boundary influence over multiple grid cells.

\subsubsection{Discrete state update}
\label{subsec:method_sim_update}

We evolve fluid velocity $\mathbf u_t:\Omega_h\rightarrow\mathbb R^3$ and (optionally) a transported
scalar $\rho_t:\Omega_h\rightarrow\mathbb R$ (which may represent a smoke or dye density) using a standard projection-style update,
implemented with differentiable discretizations of advection and spatial derivatives.
Let $\Delta t$ denote the timestep and let $\mathbf f_t$ denote an external body force field. We make use of the soft \emph{solid indicator} $\chi(\mathbf x;\theta)\in[0,1]$ given by the Gaussian-induced occupancy.

A single step consists of:
\begin{align}
\tilde{\mathbf u}_t &= \mathbf u_t + \Delta t\,\mathbf f_t,
\label{eq:sim_predict}\\
\mathbf u^{B}_t(\mathbf x) &=
\frac{\tilde{\mathbf u}_t(\mathbf x)+\Delta t\,\lambda\,\chi(\mathbf x)\,\mathbf u_{\mathrm{solid}}(\mathbf x)}
{1+\Delta t\,\lambda\,\chi(\mathbf x)},
\label{eq:brinkman_step}\\
\hat{\mathbf u}_{t+1} &= \mathcal A(\mathbf u^{B}_t),
\label{eq:sim_pressure}\\
p_t &= \mathcal P(\nabla\cdot \hat{\mathbf u}_{t+1}),
\label{eq:sim_project}\\
\rho_{t+1} &= \mathcal A_\rho(\rho_t;\mathbf u^{B}_t),
\label{eq:sim_advect_rho}\\
\mathbf u_{t+1} &= \hat{\mathbf u}_{t+1} - \nabla p_t.
\label{eq:sim_final_u}
\end{align}
Here $\mathcal A$ and $\mathcal A_\rho$ are semi-Lagrangian advection operators.
The Brinkman step~\eqref{eq:brinkman_step} relaxes velocity toward the solid velocity field
$\mathbf u_{\mathrm{solid}}$ with strength $\lambda$ in regions where $\chi$ is near one.
The pressure projection $\mathcal P(\cdot)$ is a standard incompressible projection on the full grid.

\subsubsection{Brinkman penalization for soft solid interaction}
\label{subsec:method_brinkman}

Solid-fluid interaction is handled entirely in the momentum equation via Brinkman penalization,
without modifying the incompressibility constraint or the pressure solve.

Given a soft solid indicator $\chi(\mathbf x;\theta)\in[0,1]$, we introduce a damping force density
\begin{equation}
\mathbf f_{\mathrm{Br}}(\mathbf x)
=
\lambda\,\chi(\mathbf x)\big(\mathbf u_{\mathrm{solid}}(\mathbf x)-\mathbf u(\mathbf x)\big),
\label{eq:brinkman_force_density}
\end{equation}
which relaxes fluid velocity toward the prescribed solid velocity
$\mathbf u_{\mathrm{solid}}$ in regions where $\chi$ is near one.
We apply this force using a semi-implicit
update that is stable for large $\lambda$:
\begin{equation}
\mathbf u^{B}_t(\mathbf x)
=
\frac{\tilde{\mathbf u}_t(\mathbf x)+\Delta t\,\lambda\,\chi(\mathbf x)\,\mathbf u_{\mathrm{solid}}(\mathbf x)}
{1+\Delta t\,\lambda\,\chi(\mathbf x)}.
\label{eq:brinkman_step}
\end{equation}

This formulation smoothly interpolates between fluid motion ($\chi\approx 0$) and rigid solid
motion ($\chi\approx 1$) by partially relaxing velocities toward $\mathbf u_{\mathrm{solid}}$ in the
transition region: for $0<\chi(\mathbf x)<1$,~\eqref{eq:brinkman_step} acts as a fractional drag and can
be written as $\mathbf u^B_t=(1-w)\tilde{\mathbf u}_t+w\,\mathbf u_{\mathrm{solid}}$ with
$w=\frac{\Delta t\,\lambda\,\chi}{1+\Delta t\,\lambda\,\chi}$.
The update remains differentiable with respect to the soft geometry parameters and approaches a
no-slip condition inside the solid region for large $\lambda$.
As the mask transition becomes sharper, the method increasingly approximates a sharp impermeable
boundary.

\subsubsection{Rigid-body simulation coupling moving solids}
\label{subsec:method_rigidbody}

While some tasks consider static solids (e.g., teapots fixed in space), our simulator supports the
general setting in which the solid object is a moving rigid body coupled to the surrounding fluid.
We represent rigid-body state at time $t$ by position $\mathbf x_t\in\mathbb R^3$, linear velocity
$\mathbf v_t\in\mathbb R^3$, orientation quaternion $\mathbf q_t\in\mathbb H$ (unit norm), and
world-frame angular momentum $\mathbf L_t\in\mathbb R^3$.
Given the current orientation, the corresponding world-frame inertia is
\[
\mathbf I_t^{w} = R(\mathbf q_t)\,\mathbf I_{\mathrm{body}}\,R(\mathbf q_t)^\top,
\]
and the angular velocity is recovered from
\[
\mathbf L_t = \mathbf I_t^{w}\,\boldsymbol\omega_t.
\]
The soft mask $\chi(\cdot;\theta)$ is transformed into world coordinates using the rigid-body pose,
so the induced solid region moves consistently with the object.

\paragraph{Solid velocity field}
\label{subsec:method_rigidbody_velocity}

To apply boundary conditions (Sec.~\ref{subsec:method_brinkman}) we require a solid velocity field
$\mathbf u_{\mathrm{solid}}(\mathbf x)$ on the grid.
For a rigid body, the solid velocity at a spatial location $\mathbf x$ is
\begin{equation}
\mathbf u_{\mathrm{solid}}(\mathbf x)
=
\mathbf v_t + \boldsymbol\omega_t \times (\mathbf x - \mathbf x_t),
\label{eq:rigid_velocity_field}
\end{equation}
which is evaluated at grid cell centers.

\paragraph{Hydrodynamic forces and torques from Brinkman reaction}
\label{subsec:method_rigidbody_forces}

We compute body reaction forces from the discrete momentum exchange induced by the semi-implicit
Brinkman step~\eqref{eq:brinkman_step}. Let $\tilde{\mathbf u}_t$ denote the pre-Brinkman fluid
velocity entering that step and let $\mathbf u_t^{B}$ denote the post-Brinkman velocity.
The resulting force density applied to the fluid is
\[
\mathbf f_{\mathrm{fluid}}(\mathbf x)
=
\rho_0\,\frac{\mathbf u_t^{B}(\mathbf x)-\tilde{\mathbf u}_t(\mathbf x)}{\Delta t},
\]
where $\rho_0$ is the constant fluid density used in the simulator (set to $\rho_0=1$ in our experiments).
By Newton's third law, the net force and torque on the rigid body are the negative volume sums of this
discrete force density:
\begin{align}
\mathbf f_t &=
- \sum_{\mathbf x\in\Omega_h}
\rho_0\,\frac{\mathbf u_t^{B}(\mathbf x)-\tilde{\mathbf u}_t(\mathbf x)}{\Delta t}\,h^3,
\label{eq:brinkman_force}\\
\boldsymbol\tau_t &=
- \sum_{\mathbf x\in\Omega_h} (\mathbf x-\mathbf x_t)\times
\left(
\rho_0\,\frac{\mathbf u_t^{B}(\mathbf x)-\tilde{\mathbf u}_t(\mathbf x)}{\Delta t}
\right)\,h^3.
\label{eq:brinkman_torque}
\end{align}
This estimator is differentiable with respect to $\chi(\cdot;\theta)$ through~\eqref{eq:brinkman_step}.

\paragraph{Rigid-body time integration and actuation}
\label{subsec:method_rigidbody_integration}

The translational motion is advanced with a first-order symplectic Euler step: forces are evaluated using the current state to update linear velocity explicitly, and position is then updated using the new velocity. Let $m$ denote the body mass and let
$\mathbf f_t^{\mathrm{ext}}$ denote any additional known external force applied to the body
(e.g., body gravity or thrust in the flight experiments). We update
\begin{align}
\mathbf v_{t+1} &= \mathbf v_t + \Delta t\,(\mathbf f_t + \mathbf f_t^{\mathrm{ext}})/m, &
\mathbf x_{t+1} &= \mathbf x_t + \Delta t\,\mathbf v_{t+1}.
\label{eq:rb_linear}
\end{align}

For rotation, the implementation treats the $\boldsymbol\omega$-dependent part of the Brinkman
reaction implicitly. In practice, an explicit angular update was found to be numerically unstable under the strong fluid--solid coupling regimes used in our experiments, whereas the implicit update
remained stable.
Let
\[
\mathbf r(\mathbf x)=\mathbf x-\mathbf x_t,
\qquad
\alpha(\mathbf x)=\frac{\rho_0\,\lambda\,\chi(\mathbf x)}{1+\Delta t\,\lambda\,\chi(\mathbf x)}.
\]

Using
\[
\mathbf u_{\mathrm{solid}}(\mathbf x)
=
\mathbf v_t+\boldsymbol\omega\times \mathbf r(\mathbf x),
\qquad
\mathbf r(\mathbf x)=\mathbf x-\mathbf x_t,
\]
the discrete Brinkman reaction torque can be written as
\begin{align}
\boldsymbol\tau(\boldsymbol\omega)
&=
- \sum_{\mathbf x\in\Omega_h}
\mathbf r(\mathbf x)\times
\Big(
\alpha(\mathbf x)\big(
\mathbf u_{\mathrm{solid}}(\mathbf x)-\tilde{\mathbf u}_t(\mathbf x)
\big)
\Big)\,h^3 \\
&=
- \sum_{\mathbf x\in\Omega_h}
\mathbf r(\mathbf x)\times
\Big(
\alpha(\mathbf x)\big(
\mathbf v_t+\boldsymbol\omega\times \mathbf r(\mathbf x)-\tilde{\mathbf u}_t(\mathbf x)
\big)
\Big)\,h^3 \\
\end{align}

Separating terms that depend on $\mathbf \omega$ we obtain
\begin{equation}
\begin{aligned}
\boldsymbol\tau(\boldsymbol\omega) =
&- \sum_{\mathbf x\in\Omega_h}
\mathbf r(\mathbf x)\times
\Big(
\alpha(\mathbf x)\big(
\mathbf v_t-\tilde{\mathbf u}_t(\mathbf x)
\big)
\Big)\,h^3
\\&-
\sum_{\mathbf x\in\Omega_h}
\alpha(\mathbf x)\,
\mathbf r(\mathbf x)\times
\big(
\boldsymbol\omega\times \mathbf r(\mathbf x)
\big)\,h^3.
\end{aligned}
\end{equation}
Using the vector identity
\[
\mathbf r\times(\boldsymbol\omega\times \mathbf r)
=
\big(
\|\mathbf r\|^2\mathbf I_3-\mathbf r\mathbf r^\top
\big)\boldsymbol\omega,
\]
the second term becomes
\[
\sum_{\mathbf x\in\Omega_h}
\alpha(\mathbf x)
\big(
\mathbf r(\mathbf x)\mathbf r(\mathbf x)^\top
-
\|\mathbf r(\mathbf x)\|^2\mathbf I_3
\big)\,h^3\,\boldsymbol\omega.
\]
Therefore,
\[
\boldsymbol\tau(\boldsymbol\omega)=\boldsymbol\tau_{0,t}+\mathbf K_t\,\boldsymbol\omega,
\]
where
\begin{align}
\boldsymbol\tau_{0,t}
&=
- \sum_{\mathbf x\in\Omega_h}
\mathbf r(\mathbf x)\times
\big(
\alpha(\mathbf x)\,(\mathbf v_t-\tilde{\mathbf u}_t(\mathbf x))
\big)\,h^3,\\
\mathbf K_t
&=
\sum_{\mathbf x\in\Omega_h}
\alpha(\mathbf x)
\big(
\mathbf r(\mathbf x)\mathbf r(\mathbf x)^\top
-
\|\mathbf r(\mathbf x)\|^2\mathbf I_3
\big)\,h^3.
\end{align}

We then apply an implicit Euler update to angular momentum:

\begin{align}
\mathbf L_{t+1} &= \mathbf L_t + \Delta t\,\boldsymbol\tau(\boldsymbol\omega_{t+1}),
\label{eq:rb_angmom_implicit}\\
\mathbf L_{t+1} &= \mathbf I_t^{w}\,\boldsymbol\omega_{t+1}.
\label{eq:rb_angmom_to_omega}
\end{align}
Using the decomposition
\begin{equation}
\boldsymbol\tau(\boldsymbol\omega_{t+1})
=
\boldsymbol\tau_{0,t} + \mathbf K_t\,\boldsymbol\omega_{t+1},
\label{eq:rb_tau_split}
\end{equation}
we substitute~\eqref{eq:rb_tau_split} into~\eqref{eq:rb_angmom_implicit} and use~\eqref{eq:rb_angmom_to_omega} to obtain
\begin{equation}
\mathbf I_t^{w}\,\boldsymbol\omega_{t+1}
=
\mathbf L_t + \Delta t\,\boldsymbol\tau_{0,t}
+ \Delta t\,\mathbf K_t\,\boldsymbol\omega_{t+1}.
\label{eq:rb_angmom_substitute}
\end{equation}
Rearranging terms gives the $3\times 3$ linear system
\begin{equation}
\big(\mathbf I_t^{w}-\Delta t\,\mathbf K_t\big)\,\boldsymbol\omega_{t+1}
=
\mathbf L_t + \Delta t\,\boldsymbol\tau_{0,t}.
\label{eq:rb_omega_linear_solve}
\end{equation}
After solving for $\boldsymbol\omega_{t+1}$, we update angular momentum and orientation via
\begin{align}
\mathbf L_{t+1} &= \mathbf I_t^{w}\,\boldsymbol\omega_{t+1}, \\
\mathbf q_{t+1} &=
\mathrm{Normalize}\!\left(
\Delta \mathbf q(\boldsymbol\omega_{t+1},\Delta t)\otimes \mathbf q_t
\right).
\label{eq:rb_angular}
\end{align}

This yields a $3\times 3$ linear solve per timestep and is consistent with the semi-implicit
Brinkman coupling used for the fluid velocity update.

\paragraph{Static solids as a special case}
For fixed objects, we simply set $\mathbf v_t=\mathbf 0$ and $\boldsymbol\omega_t=\mathbf 0$ for all
$t$, and keep $(\mathbf x_t,\mathbf q_t)$ constant, recovering the static-boundary simulator used
in the pouring experiments.

\subsection{Joint visual--physical optimization}
\label{subsec:method_jointopt}

We seek Gaussian parameters $\theta$ that explain the observed images while also satisfying a
simulation-driven physical objective. Using the visual reconstruction loss from
Sec.~\ref{subsec:visual_supervision} and the general physical functional from
Sec.~\ref{subsec:method_sim}, we optimize the composite objective from Equation~\ref{eq:problem_joint_objective}
\begin{equation*}
\min_{\theta}\;
\mathcal J(\theta)
=
\mathcal L_{\mathrm{vis}}(\theta)
+
\mathcal L_{\mathrm{phys}}(\theta)
+
\mathcal L_{\mathrm{reg}}(\theta),
\end{equation*}
where $\mathcal L_{\mathrm{reg}}(\theta)$ denotes optional regularization terms on the Gaussian parameters
(e.g., mild priors encouraging bounded scales or suppressing spurious isolated Gaussians).
Importantly, we do \emph{not} treat the physical term as an external post-processing step; rather,
it contributes gradients jointly with the visual term during optimization.

\paragraph{Gradient decomposition}
The gradient of~\eqref{eq:problem_joint_objective} decomposes additively as
\begin{equation}
\nabla_\theta \mathcal J(\theta)
=
\nabla_\theta \mathcal L_{\mathrm{vis}}(\theta)
+
\nabla_\theta \mathcal L_{\mathrm{phys}}(\theta)
+
\nabla_\theta \mathcal L_{\mathrm{reg}}(\theta),
\label{eq:joint_grad}
\end{equation}
where $\nabla_\theta \mathcal L_{\mathrm{vis}}$ is obtained by backpropagation through the
differentiable 3D Gaussian renderer, and $\nabla_\theta \mathcal L_{\mathrm{phys}}$ is obtained by
backpropagation through the unrolled differentiable simulator
(Secs.~\ref{subsec:method_sim}--\ref{subsec:method_rigidbody}).
Thus, the only requirement for incorporating a new physical objective is that it be expressed as a
differentiable function of the simulated trajectory (Sec.~\ref{subsec:method_sim}).

\paragraph{Balancing visual and physical objectives}
In practice, the two terms may operate on different numerical scales and may produce gradients of
very different magnitudes. To address this we introduce a tradeoff parameter for the visual and physics losses, and schedule the physical contribution during training to avoid physics gradients clashing with an intermediate visual representation.
Concretely, we implement joint optimization via a staged procedure:
(i) an initialization phase optimizing $\mathcal L_{\mathrm{vis}}$ only to obtain a visually plausible
reconstruction, followed by
(ii) a joint phase in which $\mathcal L_{\mathrm{phys}}$ is enabled and optimized together with
$\mathcal L_{\mathrm{vis}}$.
Task-specific details of the physical objective instantiations are
reported in Sec.~\ref{sec:experiments}.

\paragraph{Optimization algorithm}
We optimize~\eqref{eq:problem_joint_objective} using first-order gradient methods over the Gaussian
parameters, consistent with standard 3DGS training.
At each iteration, we (1) render images from the current Gaussians to evaluate
$\mathcal L_{\mathrm{vis}}$, (2) construct the soft mask $\chi(\cdot;\theta)$ and unroll the simulator
to evaluate $\mathcal L_{\mathrm{phys}}$, and (3) backpropagate the combined loss to update $\theta$.
Algorithm~\ref{alg:joint_opt} summarizes the procedure.

\section{Results and Discussion}
\label{sec:results}

\subsection{Experimental Setup}
\label{sec:experiments}

We evaluate \ProjectShortName on tasks that require both visual fidelity and physically meaningful behavior. Specifically, we consider two classes of problems: (1) \emph{liquid pouring}, where an object must contain and direct fluid flow through a spout, and (2) \emph{aerodynamic flight}, where an object must generate sufficient lift to maintain stable motion under gravity and a constant horizontal thrust force. The supporting dataset for the experiments in this section is available at \cite{lee2026pg3dgsdata}, and the code used to reproduce the results for \ProjectShortName is available at \cite{lee2026pg3dgssoftware}

For the pouring task, we use a dataset of teapots, where each object is observed in $20$ rendered views with fixed lighting and camera intrinsics. From these posed images alone, the method reconstructs a 3D representation of the teapot. The reconstruction is optimized to satisfy two requirements: it must match the input images, and it must permit fluid to exit through the spout under gravity in simulation. To encourage this behavior, we define a physics objective based on the mass of a tracked fluid weighted by its negative $z$-position and averaged over time, so that configurations that allow fluid to flow downward and out of the spout receive better scores. When the teapot is upright, we use the negated objective. For this task, we fix the position and orientation of the teapot throughout the simulation.

For the flight task, we use a dataset of airplanes, where each object is observed in $20$ rendered views. From these posed images, the method reconstructs a 3D representation of the airplane. In addition to matching the input views, the reconstruction is optimized to achieve favorable aerodynamic behavior in simulation. We define the physical objective as the negative of the average $z$-position over a fixed simulation horizon under gravity and a constant thrust force, using this quantity as a proxy for lift generation. For this task, we allow the position to evolve during simulation while fixing the orientation, in order to approximate a wind-tunnel-style evaluation and isolate lift effects attributable to the object’s shape in a neutral pose from those caused by uncontrolled changes in angle of attack.

We compare our method against baselines that reconstruct 3D geometry without physical supervision: standard 3D Gaussian Splatting (3DGS) for the flight task, and 3DGS, VolSDF \cite{yariv2021volume}, and NeRF for the pouring task. For the Gaussian-based methods (3DGS and \ProjectShortName), the splats are initialized from noised ground truth depth maps and trained subsequently from RGB views without depth information. The NeRF and SDF baselines are trained from the RGB views without this depth-based initialization. Consequently, cross-family comparisons are not initialization-matched and should be interpreted with this difference in mind. The cleanest controlled comparison in our experiments is therefore between 3DGS and \ProjectShortName, which share the same initialization. All methods are evaluated using the same downstream physical simulations.

Quantitatively, we report visual reconstruction quality using SSIM across views. Physical performance is evaluated using task-specific metrics: for pouring, whether fluid successfully exits the spout without leaking through the body; for flight, whether the object is able to produce lift and increase its altitude rather than falling from gravity. Unless otherwise stated, all results use the best-performing physics weight $\lambda$ selected via validation.

\subsection{Main Results: Physics-Guided Optimization}

\begin{figure}[]
\centering
\setlength{\tabcolsep}{1pt}
\setlength{\fboxsep}{0pt}

% --- CONFIGURATION ---
\newlength{\labelw}  \setlength{\labelw}{0.19\textwidth}
\newlength{\imgw}    \setlength{\imgw}{0.19\textwidth}

% --- IMAGE HELPER MACRO ---
% #1 = Filename
% #2 = Optional includegraphics arguments
\newcommand{\SqImg}[2]{%
  \fcolorbox{white}{white}{%
    \begin{minipage}[c][\imgw][c]{\imgw}
      \centering
      \includegraphics[width=\linewidth,height=\imgw,keepaspectratio,#2]{#1}
    \end{minipage}%
  }%
}

% --- TEXT LABEL MACRO ---
\newcommand{\RowLabel}[2]{%
  \parbox[c][#1][c]{\labelw}{%
    \centering
    \scriptsize \textbf{#2}%
  }%
}

\newcommand{\RefRenderCell}[1]{%
  \makebox[0pt][r]{%
    \smash{%
      \raisebox{-0.15\imgw}{\hspace{2pt}\rotatebox{90}{\scriptsize \textbf{Render}}\hspace{0pt}}%
    }%
  }%
  \fcolorbox{white}{white}{%
    \begin{minipage}[c][\imgw][c]{\imgw}
      \centering
      \begin{overpic}[width=\linewidth,height=\imgw,keepaspectratio]{#1}
        \put(42,6){%
          \makebox(0,0){%
            \fcolorbox{white}{white}{\tiny\strut Ref.\ view}%
          }%
        }
      \end{overpic}
    \end{minipage}%
  }%
}

% --- MAIN DATA BLOCK ---
\newcommand{\TeapotBlock}[1]{%
  % === ROW 1: RENDERS ===
\RefRenderCell{teapot_#1_ground_truth_image.png} &
  \SqImg{sdf_teapot_#1_render.png}{} &
  \SqImg{nerf_teapot_#1_render.png}{} &
  \SqImg{3dgs_teapot_#1_render.png}{} &
  \SqImg{physics_teapot_#1_render.png}{}\\

  % === ROW 2: CUTAWAYS ===
  \RowLabel{\imgw}{Cutaway} &
  \SqImg{sdf_teapot_#1_cutaway.png}{} &
  \SqImg{nerf_teapot_#1_cutaway.png}{} &
  \SqImg{3dgs_teapot_#1_cutaway.png}{} &
  \SqImg{physics_teapot_#1_cutaway.png}{}\\

  % === ROW 3: SIMULATIONS ===
  \RowLabel{\imgw}{Simulation} &
  \SqImg{sdf_teapot_#1_frame.png}{} &
  \SqImg{nerf_teapot_#1_frame.png}{} &
  \SqImg{3dgs_teapot_#1_frame.png}{} &
  \SqImg{physics_teapot_#1_frame.png}{} \\

  \noalign{\vspace{6pt}}
}

% ===== TABLE STRUCTURE =====
\begin{tabular}{@{} c c c c c @{}}
  \scriptsize \textbf{Output Type} &
  \scriptsize \textbf{SDF} &
  \scriptsize \textbf{NeRF} &
  \scriptsize \textbf{3DGS} &
  \scriptsize \textbf{\ProjectShortName (Ours)} \\[2pt]

  \TeapotBlock{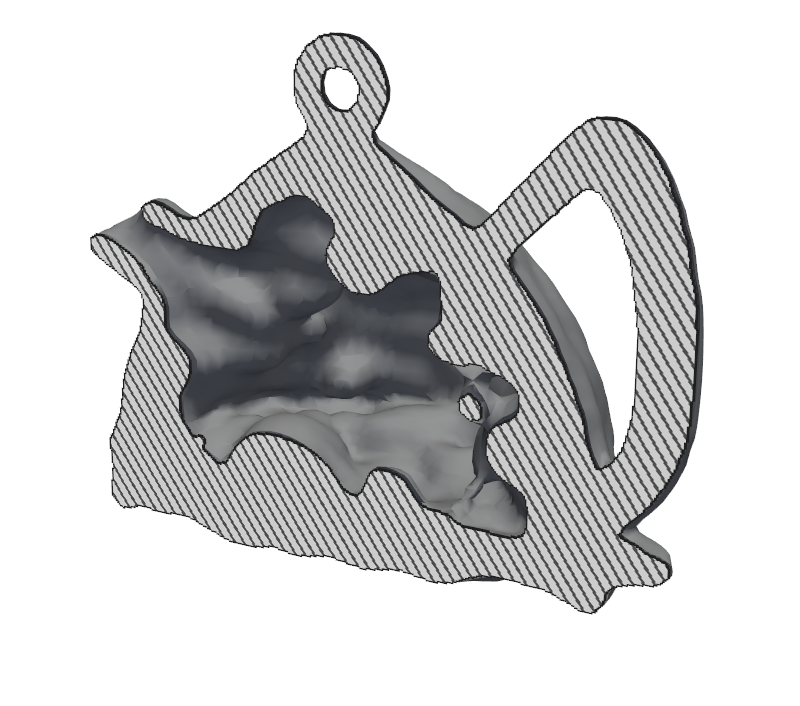}
  \TeapotBlock{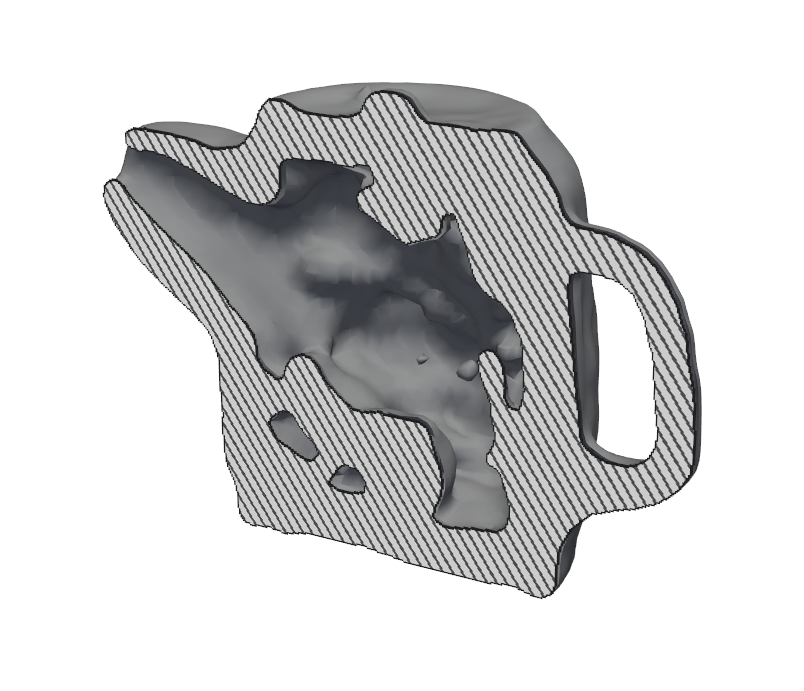}

\end{tabular}

\caption{Comparative results for two teapots. Top row: renders with ground truth view on the left. Middle row: geometry cutaways. Bottom row: simulation snapshots. Our method produces teapots with an internal cavity that allows pouring while preserving visual plausibility, whereas the baseline methods do not produce teapots that can pour.}
\label{fig:comparative_results}
\end{figure}

% --- SHARED MACROS FOR ALL FIGURES ---
\setlength{\tabcolsep}{1pt}
\setlength{\fboxsep}{0pt}
% 1. Dimensions
\newlength{\sqwidth} 
\setlength{\sqwidth}{0.19\textwidth} % 5 cols * 0.19 = 0.95 textwidth

\newcommand{\SqImg}[2]{%
  \fcolorbox{white}{white}{%
    \begin{minipage}[c][\sqwidth][c]{\sqwidth}
      \centering
      \includegraphics[width=\linewidth,height=\linewidth,keepaspectratio,#2]{#1}
    \end{minipage}%
  }%
}

\newcommand{\RenderRow}[1]{%
  % GT | Physics | 3DGS | SDF | NeRF
  \SqImg{teapot_#1_ground_truth_image.png}{} &
  \SqImg{3dgs_teapot_#1_render.png}{} &
  \SqImg{sdf_teapot_#1_render.png}{} &
  \SqImg{nerf_teapot_#1_render.png}{} &
  \SqImg{physics_teapot_#1_render.png}{} \\
  \noalign{\vspace{1pt}}
}

\newcommand{\SliceRow}[1]{%
  % GT | Physics | 3DGS | SDF | NeRF
  \SqImg{teapot_#1_ground_truth_image.png}{} &
  \reflectbox{\SqImg{sdf_teapot_#1_clean_slice.png}{}} &
  \reflectbox{\SqImg{nerf_teapot_#1_clean_slice.png}{}} &
  \reflectbox{\SqImg{3dgs_teapot_#1_clean_slice.png}{}} &
  \reflectbox{\SqImg{physics_teapot_#1_clean_slice.png}{}} \\
  \noalign{\vspace{1pt}}
}

\newcommand{\SimRow}[1]{%
  % GT | Physics | 3DGS | SDF | NeRF
  \SqImg{teapot_#1_ground_truth_image.png}{} &
  \SqImg{sdf_teapot_#1_frame.png}{} &
  \SqImg{nerf_teapot_#1_frame.png}{} &
  \SqImg{3dgs_teapot_#1_frame.png}{} &
  \SqImg{physics_teapot_#1_frame.png}{} \\
  \noalign{\vspace{1pt}}
}

\newcommand{\CutawayRow}[1]{%
  % GT | Physics | 3DGS | SDF | NeRF
  \SqImg{teapot_#1_ground_truth_image.png}{} &
  \SqImg{sdf_teapot_#1_cutaway.png}{} &
  \SqImg{nerf_teapot_#1_cutaway.png}{} &
  \SqImg{3dgs_teapot_#1_cutaway.png}{} &
  \SqImg{physics_teapot_#1_cutaway.png}{} \\
  \noalign{\vspace{1pt}}
}

\begin{figure}[p] % 'p' forces it to a dedicated page
\centering
\begin{tabular}{@{} c c c c c @{}}
\scriptsize \textbf{\shortstack{Ground\\Truth}} &
  \scriptsize \textbf{\shortstack{SDF\\Render}} &
  \scriptsize \textbf{\shortstack{NeRF\\Render}} &
  \scriptsize \textbf{\shortstack{3DGS\\Render}} &
  \scriptsize \textbf{\shortstack{\ProjectShortName\\(Ours) Render}} \\[1pt]
  
  % \RenderRow{00}
  \RenderRow{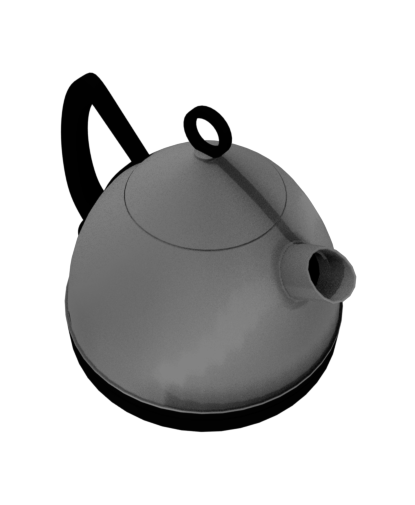}
  \RenderRow{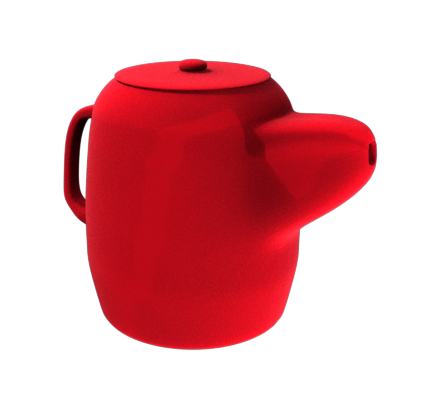}
  \RenderRow{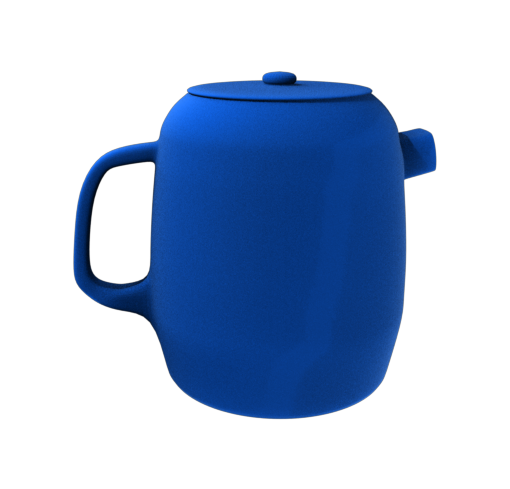}
  \RenderRow{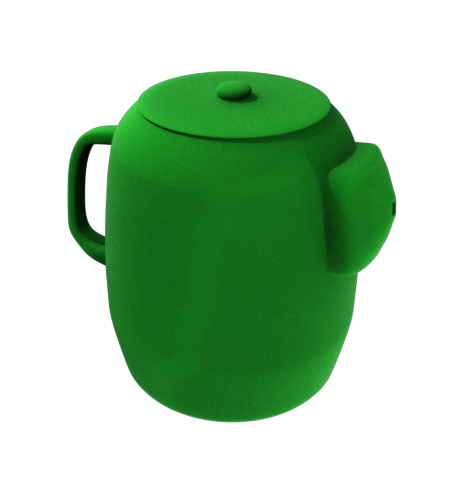}
  % Split here if the page is too full
  \RenderRow{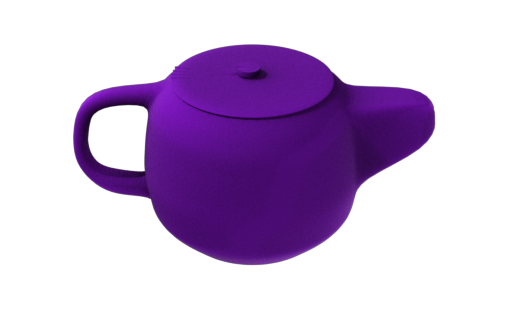}
  \RenderRow{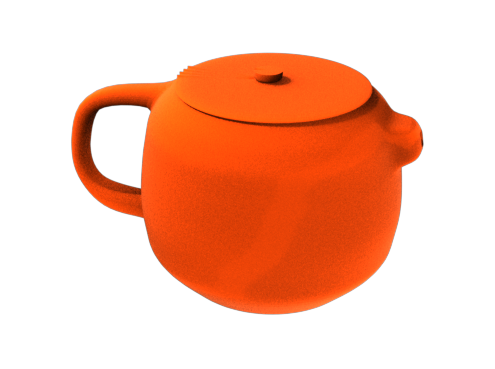}
  % \RenderRow{07}
  % \RenderRow{08}
  % \RenderRow{09}
  % \RenderRow{10}
  % \RenderRow{11}
  % \RenderRow{12}
  % \RenderRow{13}
  % \RenderRow{14}

\end{tabular}
\caption{Full Render Comparison (Teapots 01--06). All methods are rendered from the test viewpoint. We can see from this image that our method produces visually comparable results to those of the baseline, indicating that our physics objective does not substantially degrade visual quality.}
\label{fig:render_results}
\end{figure}

\begin{figure}[p]
\centering
\begin{tabular}{@{} c c c c c @{}}
\scriptsize \textbf{\shortstack{Reference\\View}} &
\scriptsize \textbf{\shortstack{SDF\\Cutaway}} &
\scriptsize \textbf{\shortstack{NeRF\\Cutaway}} &
\scriptsize \textbf{\shortstack{3DGS\\Cutaway}} &
\scriptsize \textbf{\shortstack{\ProjectShortName\\(Ours) Cutaway}} \\[1pt]
  
  % \SliceRow{00}
  \CutawayRow{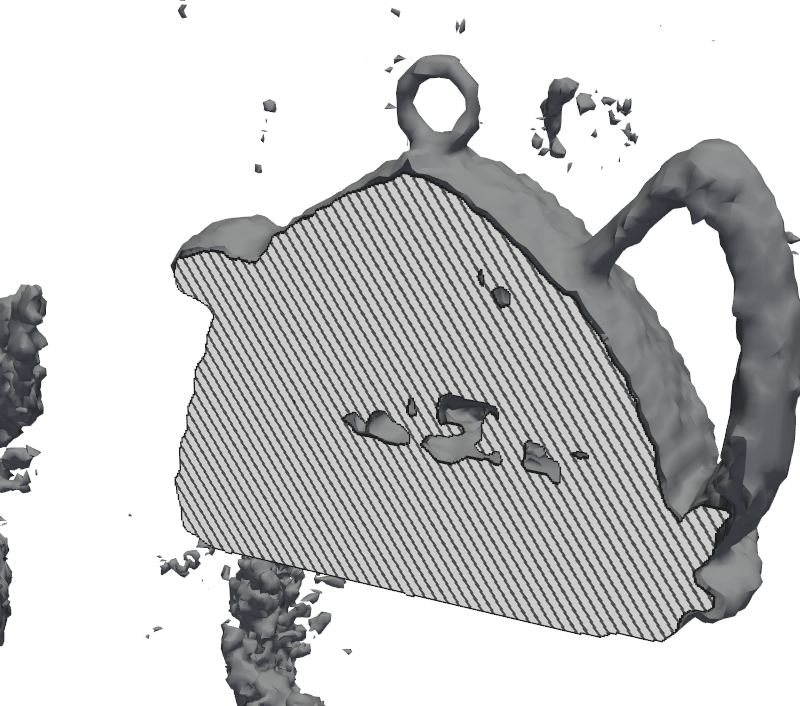}
  \CutawayRow{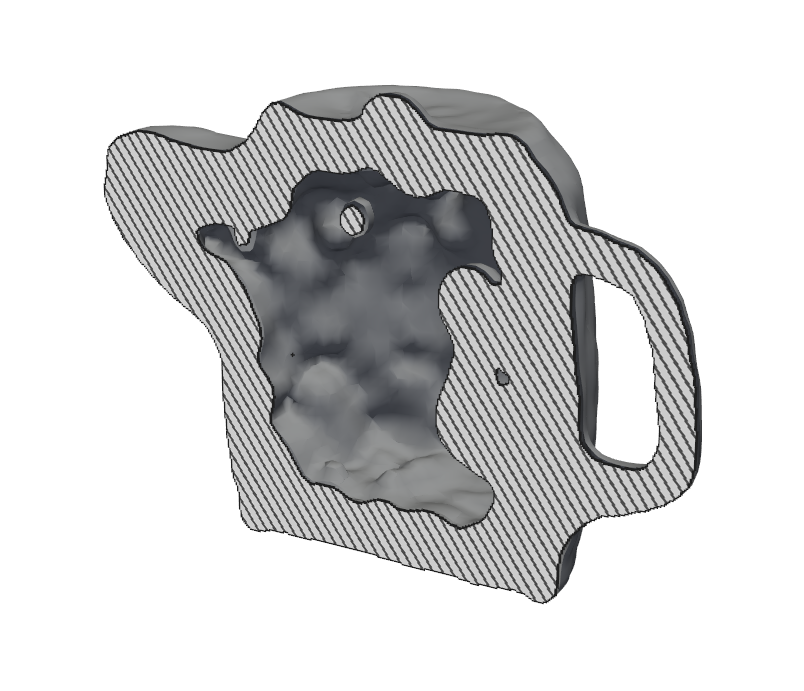}
  \CutawayRow{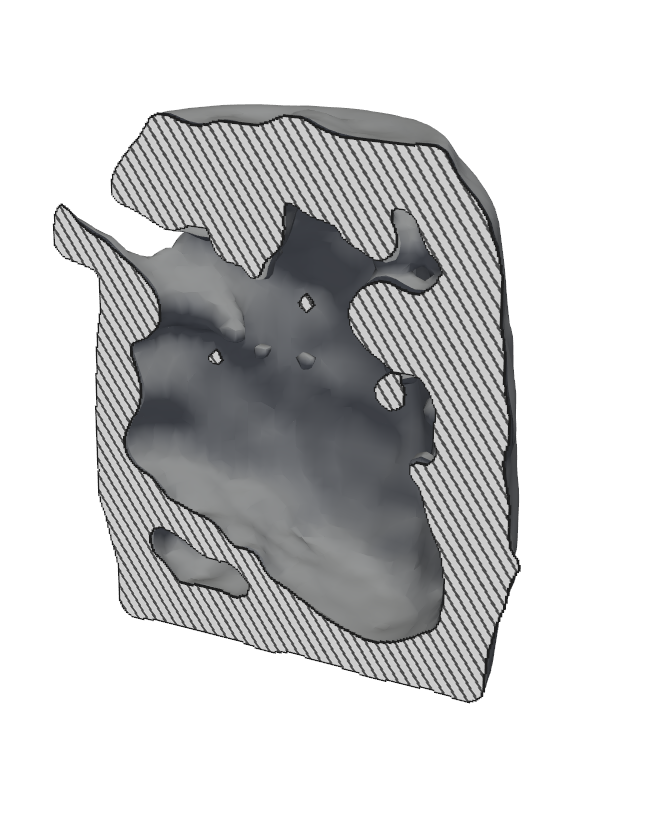}
  \CutawayRow{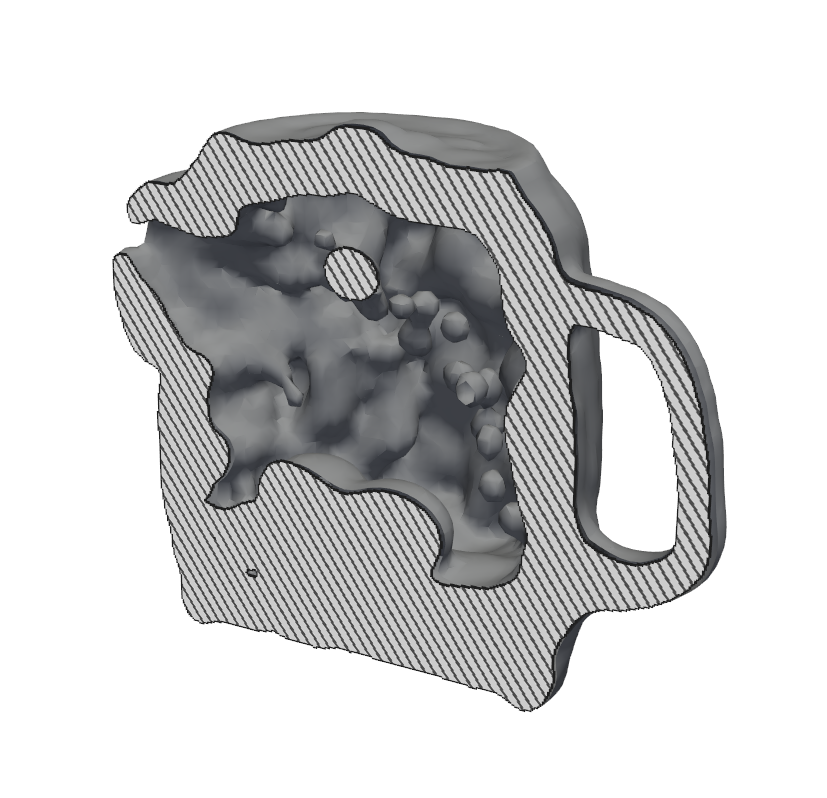}
  \CutawayRow{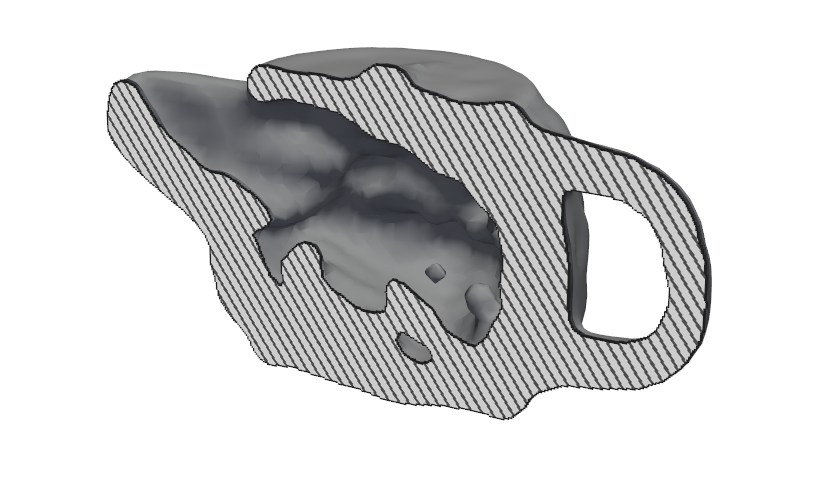}
  \CutawayRow{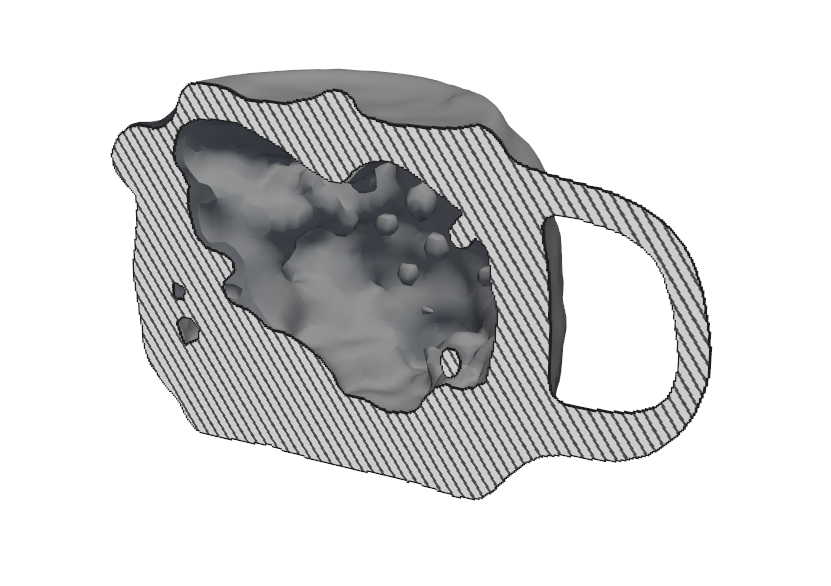}
  % \SliceRow{07}
  % \SliceRow{08}
  % \SliceRow{09}
  % \SliceRow{10}
  % \SliceRow{11}
  % \SliceRow{12}
  % \SliceRow{13}
  % \SliceRow{14}

\end{tabular}
\caption{Geometry cutaway Comparison (Teapots 01--06). The first column is the reference image; subsequent columns show the internal geometry extracted from the density field. Note that our method produces a clear path from an internal cavity to the outside of the object.}
\label{fig:cross_section_results}
\end{figure}

\begin{figure}[p]
\centering
\begin{tabular}{@{} c c c c c @{}}
\scriptsize \textbf{\shortstack{Reference\\View}} &
\scriptsize \textbf{\shortstack{SDF\\Simulation}} &
\scriptsize \textbf{\shortstack{NeRF\\Simulation}} &
\scriptsize \textbf{\shortstack{3DGS\\Simulation}} &
\scriptsize \textbf{\shortstack{\ProjectShortName (Ours)\\Simulation}} \\[1pt]
  % \SimRow{00}
  \SimRow{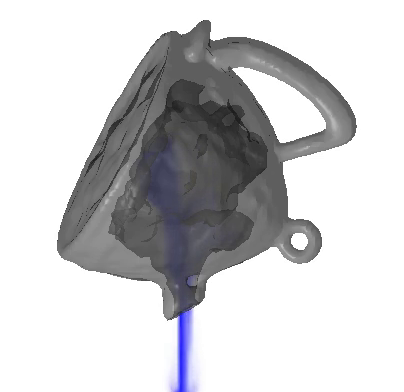}
  \SimRow{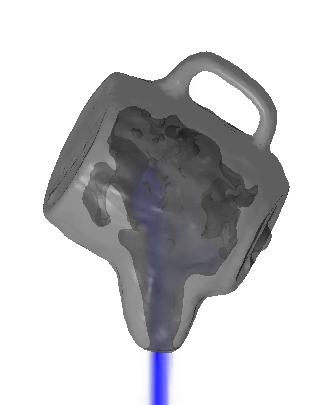}
  \SimRow{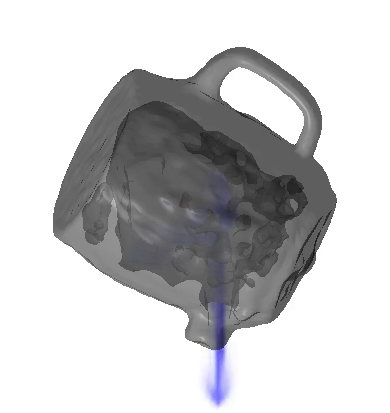}
  \SimRow{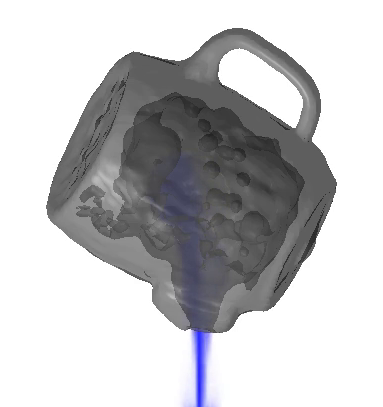}
  \SimRow{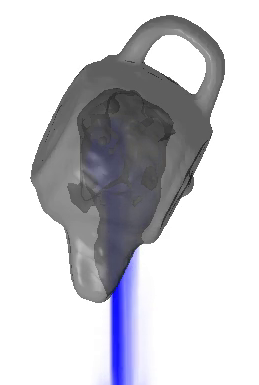}
  \SimRow{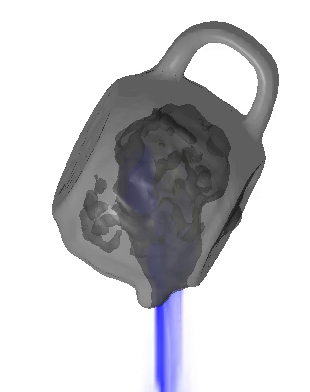}
  % \SimRow{07}
  % \SimRow{08}
  % \SimRow{09}
  % \SimRow{10}
  % \SimRow{11}
  % \SimRow{12}
  % \SimRow{13}
  % \SimRow{14}

\end{tabular}
\caption{Physics Simulation Comparison (Teapots 01--06). Snapshot taken at final frame of each simulation. The first column is the static reference image. We can see that our method allows fluid (blue) to be poured out from the teapot, while baselines trap it inside.}
\label{fig:simulation_results}
\end{figure}

\begin{figure}[]
    \centering
    \setlength{\tabcolsep}{2pt}
    \setlength{\fboxsep}{0pt}
    \setlength{\fboxrule}{0.4pt}

    \includegraphics[width=\textwidth]{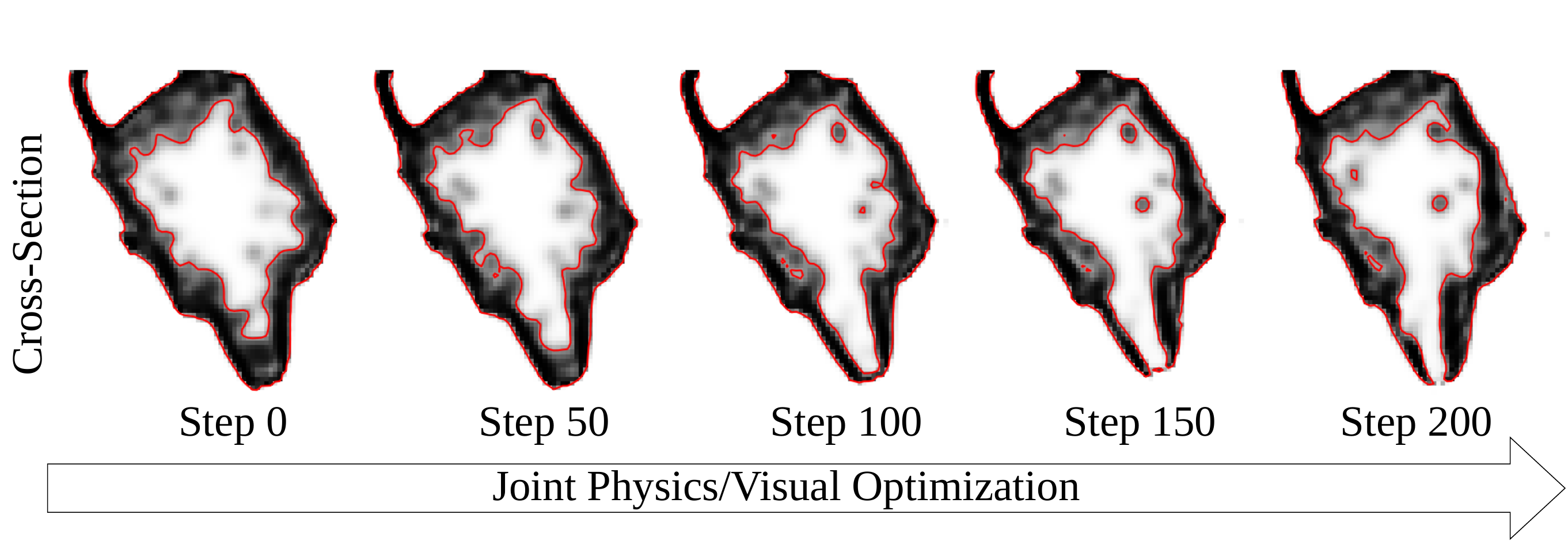}

    \caption{Training progression for Teapot 05.  
    Midplane cross-section views are shown at different training iterations, illustrating how the physics gradient shapes the spout for fluid to successfully pour out.}
    \label{fig:progression_teapot05}
\end{figure}

\begin{figure}
    \centering
    \includegraphics[width=\linewidth]{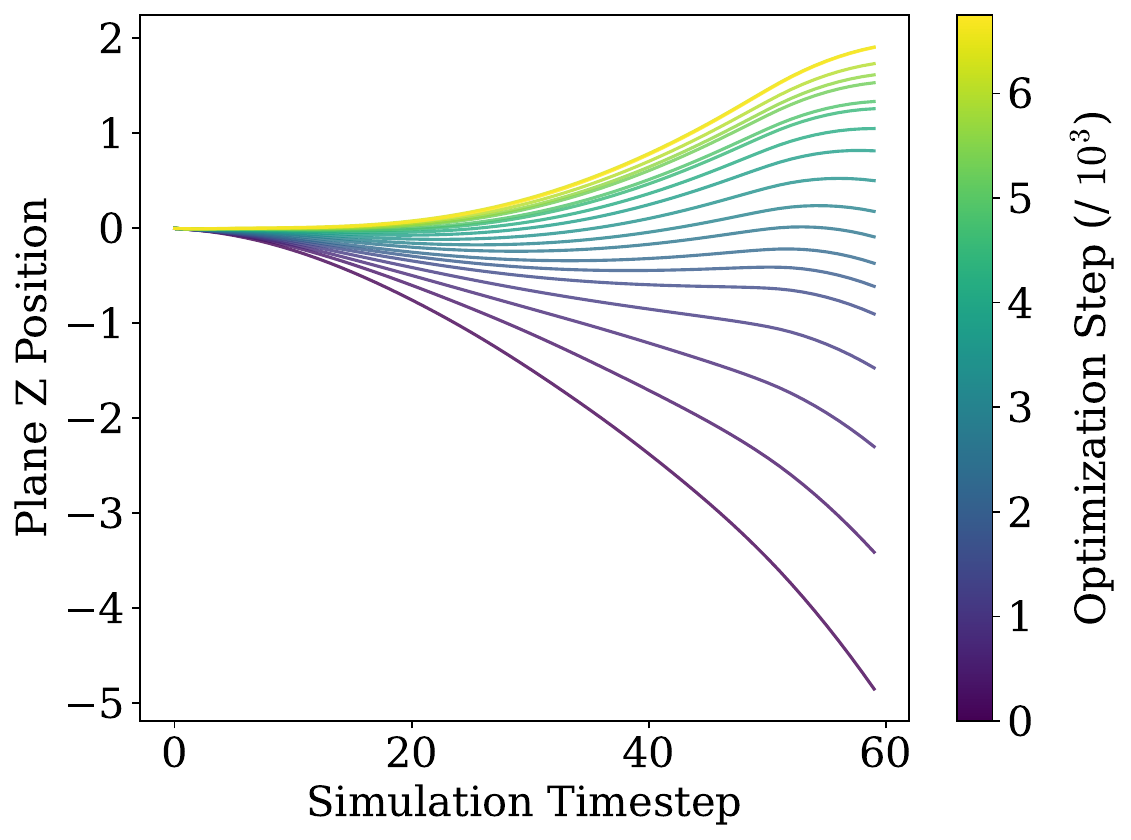}
    \caption{Trajectories for paper plane over optimization process. The z position of the plane is shown over each timestep and plotted for multiple optimization steps according to color, illustrating the increase in lift production over the training process.}
    \label{fig:plane_trajectories_optimization}
\end{figure}

\begin{figure}[]
    \centering
    \includegraphics[width=\textwidth]{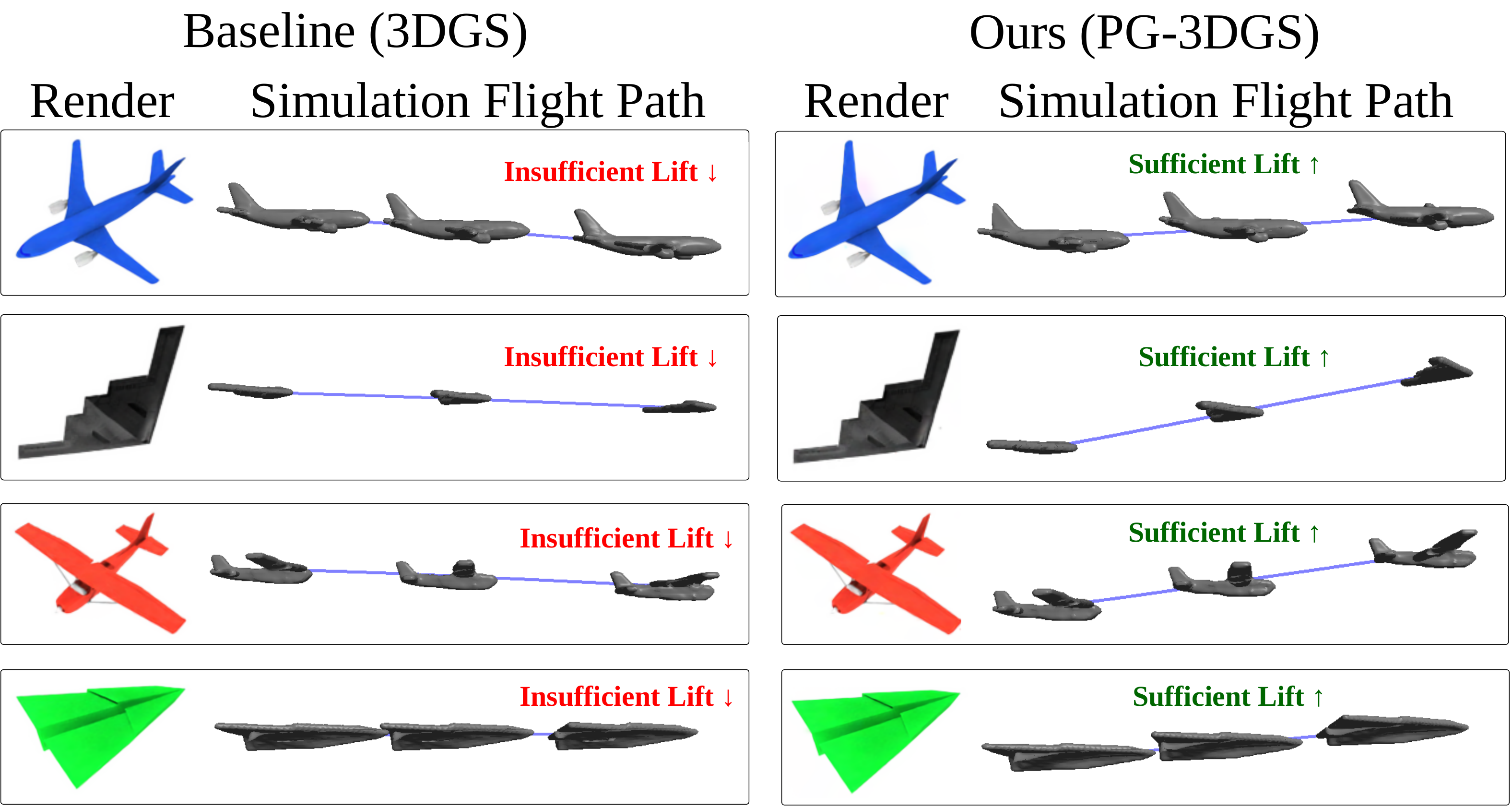}

  \caption{Comparison of airplane motion between the baseline (3DGS) and our method.  
  Baseline methods fail to produce sufficient lift under gravity and a thrust force, causing the airplanes to gradually fall. 
  By incorporating a physics-based loss, our method produces stable and realistic flight paths, allowing the airplanes to gain altitude throughout the simulation.}
    \label{fig:plane_comparison}
\end{figure}

Figures~\ref{fig:comparative_results}--\ref{fig:plane_comparison} summarize the primary qualitative results. Across all tasks, \ProjectShortName reconstructs objects that are visually comparable to appearance-only baselines while exhibiting substantially improved physical behavior.

For the pouring task, Fig.~\ref{fig:comparative_results} shows renders, cutaways, and simulation snapshots for representative teapots. While all methods produce visually plausible external views, only \ProjectShortName consistently recovers internal cavities and connected spouts that permit fluid flow within the recovered geometry. Baseline methods tend to seal the interior or introduce disconnected voids, causing fluid to pool or reamain trapped during simulation. The NeRF baseline additionally tends to place spurious mass away from the object; although this has limited effect on image rendering, it becomes apparent in cutaway views and distorts the simulated dynamics.

These effects are more clearly visible in the cutaway comparisons in Fig.~\ref{fig:cross_section_results}. \ProjectShortName recovers coherent hollow interiors aligned with the functional structure of the teapot, whereas the baselines often collapse or fragment the interior despite matching the external appearance. As a result, the downstream simulations in Fig.~\ref{fig:simulation_results} show successful pouring only for our method.

Importantly, incorporating the physics objective does not significantly degrade visual quality. Fig.~\ref{fig:render_results} demonstrates that rendered images from \ProjectShortName remain visually indistinguishable from those of 3DGS and NeRF under the same viewpoints.

Additional results for the teapots are shown in \ref{app:teapot_results}. Results for renders can be found in Figures~\ref{fig:render_results_appendix_1} and \ref{fig:render_results_appendix_2}, cutaways can be be found in \ref{fig:cross_section_results_appendix_1} and \ref{fig:cross_section_results_appendix_2} and simulation results can be found in \ref{fig:simulation_results_appendix_1}, and \ref{fig:simulation_results_appendix_2}.

The training dynamics are illustrated in Fig.~\ref{fig:progression_teapot05}. Early iterations resemble standard appearance-driven reconstruction. As optimization proceeds, the physics gradient reshapes the geometry to open and align the spout while preserving external appearance. This behavior is difficult to achieve using appearance losses alone.

We observe similar trends in the flight task. As shown in Fig.~\ref{fig:plane_comparison}, airplanes reconstructed using baseline methods gradually lose altitude. In contrast, \ProjectShortName consistently produces geometries that generate sufficient lift to sustain flight over the simulation horizon. These results indicate that the physics loss induces meaningful geometric changes.

We show training dynamics for a paper plane experiment in Fig.~\ref{fig:plane_trajectories_optimization}, by plotting the z position over time for multiple points in the optimization process. We notice a steady change from downward to upward trajectories as training progresses.

\subsection{Benchtop Physical Lift Validation}
\label{sec:real_lift_validation}

\begin{figure}[H]
\includegraphics[width=\textwidth]{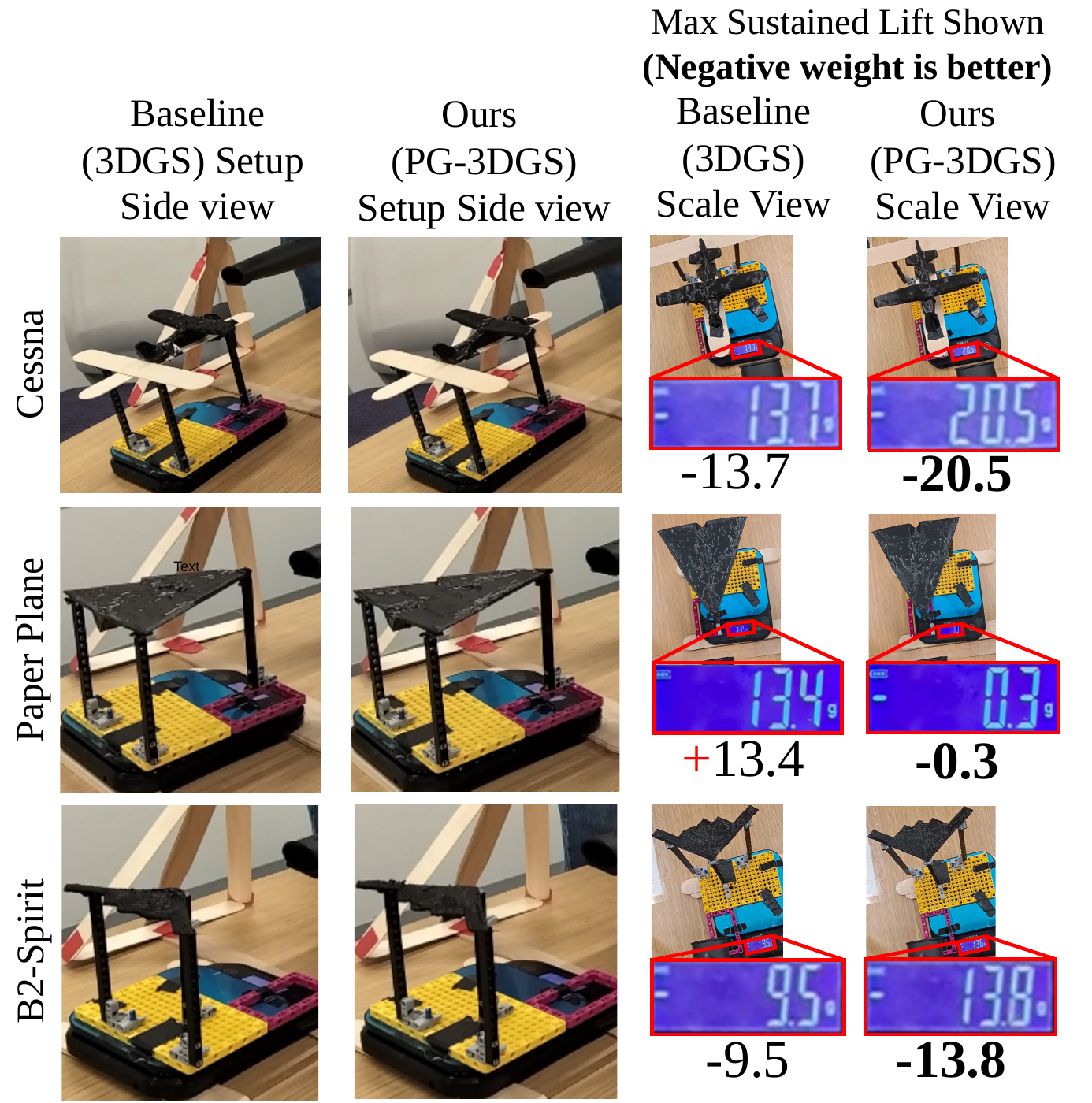}
\caption{Benchtop physical lift tests under identical blower airflow. Each row corresponds to one aircraft shape, and columns show synchronized side/top views for the appearance-only baseline and \ProjectShortName. Across all three shapes, \ProjectShortName yields higher scale-measured lift (Table~\ref{tab:physical_lift}). (Note that more negative scale readings indicate more lift.)} 
\label{fig:physical_lift_bench}
\end{figure}

To assess whether the simulation-driven lift improvements translate to the physical world, we performed benchtop lift tests using 3D-printed aircraft generated by our method and an appearance-only baseline. We tested three shapes (Cessna, B-2 Spirit, and paper plane) under the same airflow setup using a directed blower and measured the resulting change in scale reading as a proxy for upward lift force. For each shape, the test geometry, blower placement, and measurement procedure were kept fixed across methods.

Table~\ref{tab:physical_lift} summarizes the measured lift values. Across all three aircraft, \ProjectShortName produced higher lift than the baseline under identical test conditions, consistent with the trends observed in simulation. Figure~\ref{fig:physical_lift_bench} shows representative synchronized side-view and top-view frames from the physical tests. Videos of the benchtop lift experiments are available with the accompanying project materials. For each aircraft, the side-view and top-view recordings are provided, with ours and the 3DGS baseline tested sequentially for each plane.

These benchtop experiments are intended as a real-world proxy validation rather than a controlled wind-tunnel study; nevertheless, the consistent improvement across three distinct aircraft shapes provides evidence that physics-guided optimization induces geometric changes that remain effective beyond the simulator.

\begin{table}[]
\centering
\caption{Benchtop physical lift validation under identical blower airflow. Note that we report the upward force in this table, which is the negation of the scale reading. 
Lift is computed from the maximum sustained change in lift, meaning after the blower has started and the scale reading has reached a steady state (higher is better). 
\ProjectShortName outperforms the appearance-only baseline on all three aircraft shapes.}
\label{tab:physical_lift}
\setlength{\tabcolsep}{6pt}
\begin{tabular}{lcccc}
\toprule
Shape & Baseline Lift (grams) & Ours Lift (grams) & $\Delta$ Lift (grams) \\
\midrule
Cessna     & 13.7 & \textbf{20.5} & 6.8  \\
Paper plane& -13.4 & \textbf{0.3} & 13.7\\
B-2 Spirit & 9.5 & \textbf{13.8} & 4.3  \\
\midrule
Mean       & 3.27 & \textbf{11.53} & 8.27 \\
\bottomrule
\end{tabular}
\end{table}

\subsection{Ablation Study}

\begin{figure}
    \centering
    \includegraphics[width=1.0\linewidth, trim={0 0 0 0pt}, clip]{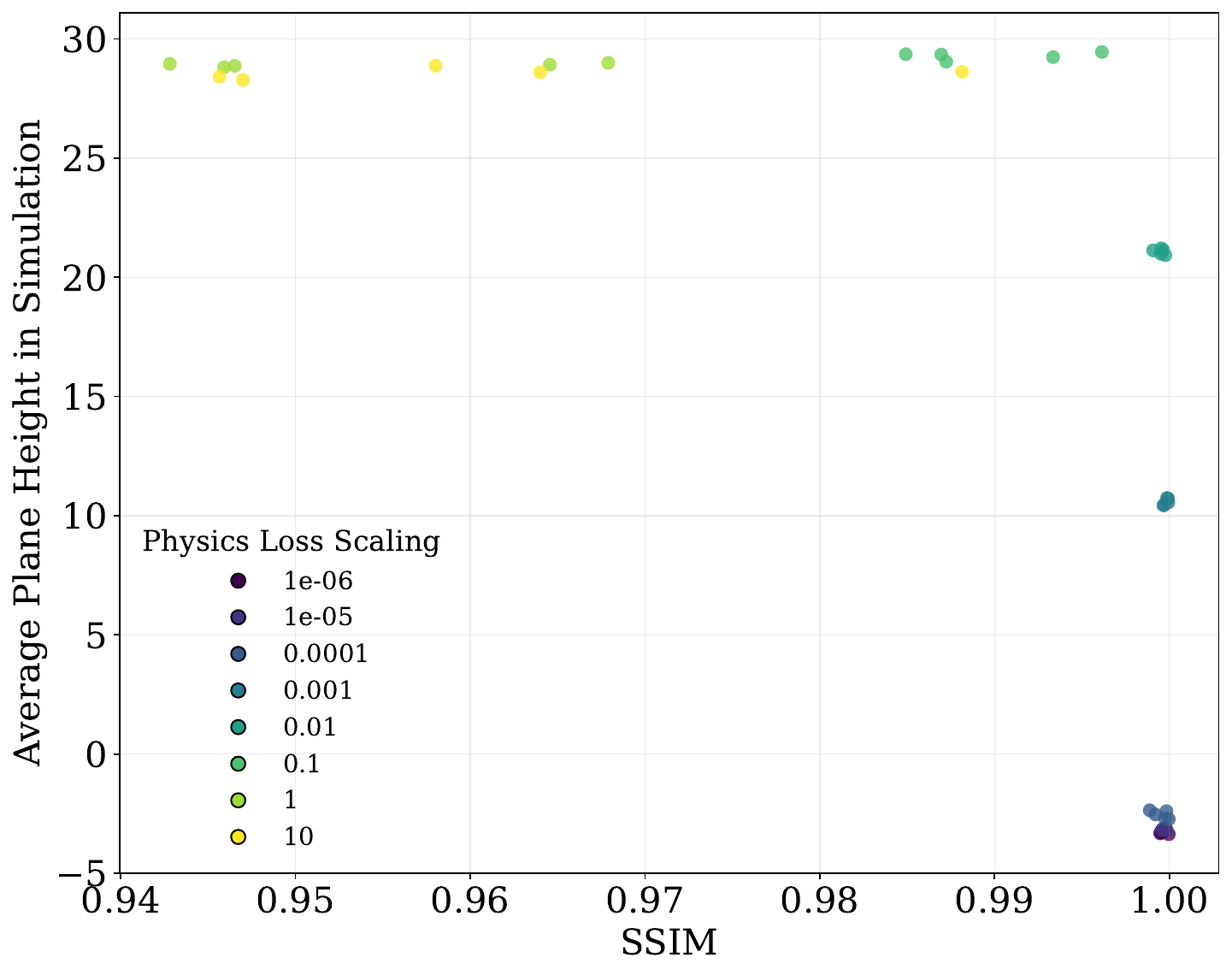}
    \caption{This plot shows that while the satisfaction of the physics objective does increase meaningfully as one increases the loss scaling, the average SSIM across rendered viewpoints does not decrease meaningfully until very high scalings are applied. This suggests that the satisfaction of physical functionality does not need to sacrifice visual appearance.}
    \label{fig:physics_weight_ablation}
\end{figure}

We evaluate the effect of the physics-loss strength by sweeping the physics weight across 8 distinct values spanning several orders of magnitude.
For each weight, we train the model under identical initialization and optimization
settings and record the final checkpoint performance over 5 experiments.
We report (i) average SSIM on rendered views as the visual metric and
(ii) the physics objective value, defined as the negative physics loss
(i.e., higher is better). For this study, we evaluate a plane (B2-Spirit) given a constant thrust force and no rotation, with a simple loss on the negative of the average z position, encouraging planes that are able to produce lift.

Figure~\ref{fig:physics_weight_ablation} plots SSIM versus the physics objective for all 40 runs,
with color indicating the corresponding physics weight.
Several consistent trends emerge.

First, increasing $\lambda$ from zero yields a substantial improvement in the
physics objective while SSIM remains nearly constant (or no worse than lower $\lambda$ trials) over a broad range of weights, until around $\lambda=0.1$.
This indicates that physics supervision primarily resolves geometric ambiguities
that are not constrained by image formation, rather than introducing a strong
appearance–function trade-off.

Second, for very small $\lambda$ where the model behaves similarly to
appearance-only 3DGS: SSIM is high, but the physics objective remains poor.
As $\lambda$ increases, the solution moves toward geometries that satisfy
the physical objective without a noticeable drop in visual quality.

Finally, at the largest weights in the sweep, we observe diminishing returns
in the physics objective and mild degradation in SSIM.
In this regime, the optimization becomes increasingly dominated by the physics term,
which can pull geometry slightly away from the visually optimal basin.

Overall, the sweep demonstrates the existence of a wide interval of physics weights
that substantially improve functional performance while preserving visual fidelity.
This supports the interpretation of \ProjectShortName as using physics guidance
to select among visually plausible geometries those that are physically functional,
rather than sacrificing appearance to enforce physical constraints.

\section{Conclusion}
\label{sec:conclusion}

This work argues that appearance alone is an incomplete supervision signal for 3D structure generation when the resulting assets are meant to be used rather than only rendered. Standard inverse-graphics objectives admit many geometries that are visually consistent under different views yet physically nonfunctional, because the image formation process is insensitive to internal structure, watertightness, connectivity, and other properties that govern physical behavior. 

\ProjectShortName ~addresses this ambiguity by integrating differentiable physics as an additional source of evidence during optimization, producing reconstructions that remain visually faithful while satisfying downstream physical objectives.
One of our core contributions is a practical coupling between an explicit 3D Gaussian representation and grid-based physics through a differentiable soft occupancy mask. The mask provides a smooth interface from Gaussian parameters to simulator coefficients, enabling gradients from physical objectives to backpropagate to Gaussian means, covariances, and opacities. Empirically, this coupling is sufficient to drive targeted, interpretable geometric changes that are hard to obtain from image losses alone. In the pouring task, the dominant failure mode for appearance-only baselines is an interior that is sealed from the spout opening despite a plausible exterior. \ProjectShortName resolves this by reshaping interior occupancy to recover a connected cavity and an opening at the spout, which directly translates into successful flow in simulation. In the airplane lift task, the physics objective induces modifications that increase aerodynamic effectiveness (e.g., changes to wing thickness/camber and local surface orientation), yielding sustained flight trajectories under identical simulation conditions. Across both tasks, the key observation is that physics gradients select among visually plausible solutions and favor those with functional structures.

Our second observation is that the physics guidance can be introduced without a large visual penalty when applied in a staged manner. Initializing with a purely visual phase produces a good appearance prior over Gaussians, after which the physics term acts as a refinement signal that resolves physically relevant ambiguities. The ablation trend (physics weight versus visual metrics) is consistent with the interpretation that the solution set contains a broad visually acceptable basin, and physics provides a direction within that basin toward functional configurations. In this sense, our method behaves like a form of posterior inference in which physical evidence disambiguates underdetermined geometry, rather than as a competing objective that necessarily degrades rendering quality.

The benchtop lift validation provides an important check on whether simulation-driven improvements reflect meaningful geometry changes rather than simulator-specific artifacts. While the setup is not a wind-tunnel study, the consistent improvement across three distinct aircraft shapes under identical airflow conditions suggests that the optimized geometries capture real aerodynamic effects. This result is encouraging given the multiple sources of mismatch between simulation and reality (unknown turbulence, Reynolds-number differences, printer tolerances, surface roughness, and alignment variability). More broadly, it supports the central premise that incorporating physics during reconstruction can improve downstream utility beyond the training simulator.

There are several limitations that shape when \ProjectShortName is most effective. First, the physical objective is only as reliable as the simulator and its boundary treatment; the soft-mask Brinkman coupling trades boundary sharpness for differentiability, and overly smooth masks can blur fine geometric features that matter for certain regimes, while overly sharp masks can yield poorly conditioned gradients. Second, grid resolution and timestep size constrain the fidelity of the simulated signal, especially for thin structures, narrow channels, or boundary-layer effects relevant to aerodynamics. Third, the optimization is task-conditional: the geometry changes are those that improve the specified objective, which may not coincide with the true physical shape of the object if the objective is misspecified or incomplete. Finally, unrolling differentiable simulation is computationally expensive and introduces additional hyperparameters (physics weight schedules, simulation horizon, and loss frequency) that can affect convergence.

These limitations point to several directions for future work. On the simulation side, higher-fidelity or multi-resolution solvers, improved immersed-boundary treatments, and better handling of high-Reynolds-number flows could expand the range of physical phenomena that can be reliably optimized. On the learning side, automatic balancing strategies (e.g., adaptive gradient normalization across visual and physics terms) could reduce tuning and improve robustness across tasks. On the representation side, extending the Gaussian-to-physics interface to incorporate additional physical properties—such as spatially varying material parameters, compliance, or permeability—would enable inverse design beyond rigid solids. Finally, combining simulation-based supervision with real-world measurements in a closed loop (e.g., small numbers of physical tests to calibrate or correct simulation bias) is a promising route toward reconstruction pipelines that produce assets that both render well and behave correctly in deployment.

\section*{CRediT authorship contribution statement}

\textbf{Zachary Lee:} Conceptualization, Methodology, Software, Validation, Formal analysis, Investigation, Data curation, Visualization, Writing -- original draft, Writing -- review \& editing. \textbf{Maxwell Jacobson:} Software, Writing -- review \& editing. \textbf{Yexiang Xue:} Conceptualization, Methodology, Supervision, Funding acquisition, Writing -- review \& editing.

\section*{Acknowledgements}
% Describe financial support and the role of the funding source, if any.
This research has been supported by NSF Career Award IIS-2339844, DOE – Fusion Energy
Science grant: DE-SC0024583.

\section*{Declaration of competing interest}
% State any competing interests, or explicitly state that none exist.
Zachary Lee reports financial support was provided by National Science Foundation. Maxwell Jacobson reports was provided by US Department of Energy. Yexiang Xue reports financial support was provided by US Department of Energy. Yexiang Xue reports financial support was provided by National Science Foundation. If there are other authors, they declare that they have no known competing financial interests or personal relationships that could have appeared to influence the work reported in this paper. 

\section*{Declaration of generative AI and AI-assisted technologies in the writing process}
% Add declaration here if you used AI tools in the writing process.
During the preparation of this work, the authors used ChatGPT in order to improve the clarity and presentation of the manuscript. After using this tool, the authors reviewed and edited the content as needed and take full responsibility for the content of the manuscript.

\section*{Data availability}
% Provide a brief statement on data/code availability.
The software and supporting dataset used in this study are publicly available. The PG-3DGS software release is archived at Zenodo, \url{https://doi.org/10.5281/zenodo.19392335}. The supporting dataset, \emph{PG-3DGS Supporting Dataset: Multiview Teapot and Airplane Images, Depth Maps, Masks, and Camera Parameters}, is available at Mendeley Data, \url{https://doi.org/10.17632/xz7tkg2zhd.1}.

% ================================================================
% References
% ================================================================
\bibliographystyle{elsarticle-num}  % or elsarticle-num-names / elsarticle-harv
\bibliography{main}

\appendix
\input{appendix}

\end{document}

%% file: method_flowchart_raw.tex
\newlength{\FigW}
\setlength{\FigW}{0.96\linewidth}
\begin{tikzpicture}[
  % lock font metrics so it doesn't depend on document class
  font=\fontsize{10pt}{12pt}\selectfont,
  >={Stealth[length=2.2mm]},
  % --- Color palette ---
  viscol/.style={draw=blue!65!black},
  physcol/.style={draw=teal!60!black},
  neutralcol/.style={draw=black!70},
  % --- Blocks (NO \textwidth/\linewidth here) ---
  block/.style={
    draw, rounded corners=2pt, thick,
    align=center,
    text width=0.35\FigW,
    minimum height=8mm,
    inner sep=3pt,
    fill=white
  },
  centerblock/.style={
    draw, rounded corners=2pt, thick,
    align=center,
    text width=0.70\FigW,
    minimum height=8mm,
    inner sep=3pt,
    fill=white
  },
  tinyblock/.style={
    draw, rounded corners=2pt, thick,
    align=center,
    text width=0.36\FigW,
    minimum height=7mm,
    inner sep=3pt,
    fill=white
  },
  inputblock/.style={
    draw, rounded corners=2pt, thick,
    align=left,
    text width=0.34\FigW,
    minimum height=7mm,
    inner sep=3pt,
    fill=white
  },
  % semantic variants
  param/.style={centerblock, neutralcol, fill=black!3},
  visstep/.style={block, viscol, fill=blue!4},
  physstep/.style={block, physcol, fill=teal!4},
  lossvis/.style={block, viscol, fill=blue!10},
  lossphys/.style={block, physcol, fill=teal!10},
  merge/.style={centerblock, neutralcol, fill=black!3},
  update/.style={centerblock, neutralcol, fill=black!1},
  line/.style={->, thick, draw=black!70},
  visline/.style={->, thick, draw=blue!65!black},
  physline/.style={->, thick, draw=teal!60!black},
  dashline/.style={->, thick, dashed, draw=black!60},
  lanevisstyle/.style={draw=blue!65!black, very thick, rounded corners=3pt, inner sep=6pt, fill=blue!2},
  lanephysstyle/.style={draw=teal!60!black, very thick, rounded corners=3pt, inner sep=6pt, fill=teal!2},
  % inputs with colored bar
  inputvis/.style={
    inputblock,
    viscol, fill=blue!3,
    inner xsep=3mm,
    path picture={
      \fill[blue!55!black]
        (path picture bounding box.south west)
        rectangle ([xshift=2mm]path picture bounding box.north west);
    }
  },
  inputphys/.style={
    inputblock,
    physcol, fill=teal!3,
    inner xsep=3mm,
    path picture={
      \fill[teal!55!black]
        (path picture bounding box.south west)
        rectangle ([xshift=2mm]path picture bounding box.north west);
    }
  },
  % gradient arrows
  gradvis/.style={->, thick, densely dashed, draw=blue!65!black, shorten <=2pt, shorten >=2pt},
  gradphys/.style={->, thick, densely dashed, draw=teal!60!black, shorten <=2pt, shorten >=2pt},
  gradline/.style={->, thick, densely dashed, draw=black!60, shorten <=2pt, shorten >=2pt},
  gradlabel/.style={fill=white, inner sep=1pt, font=\fontsize{8pt}{10pt}\selectfont},
  column sep=14mm, row sep=6mm,
]
% tcolorbox styles used by \tcbox
\tcbset{
  vislossbox/.style={
    on line, boxrule=0.6pt,
    colback=blue!10, colframe=blue!65!black,
    arc=2pt, boxsep=1pt,
    left=2pt,right=2pt,top=1pt,bottom=1pt
  },
  physlossbox/.style={
    on line, boxrule=0.6pt,
    colback=teal!10, colframe=teal!60!black,
    arc=2pt, boxsep=1pt,
    left=2pt,right=2pt,top=1pt,bottom=1pt
  },
  philossbox/.style={
    on line, boxrule=0.6pt,
    colback=purple!10, colframe=purple!70!black,
    arc=2pt, boxsep=1pt,
    left=2pt,right=2pt,top=1pt,bottom=1pt
  }
}

\node[param] (theta)
{3D Gaussian parameters $\theta=\{(\mu_i,\Sigma_i,\alpha_i,c_i)\}_{i=1}^N$};

% --- Two-column branch layout ---
\matrix (M) [below=15mm of theta, matrix of nodes, nodes in empty cells] {
  \node[inputvis] (imgin)
  {\faImages\ \textbf{Inputs}\\
   Calibrated images\\ $\mathcal I=\{I_k\}_{k=1}^K$,
   cameras $\{\pi_k\}$}; &
  \node[inputphys] (desin)
  {\faTasks\ \textbf{Input}\\
   Desired physical behavior\\
   objective \tcbox[philossbox]{$\Phi(\cdot)$}}; \\
  \node[visstep]  (render) {Differentiable renderer\\ $\hat I_k=\mathcal R(\theta;\pi_k)$}; &
  \node[physstep] (mask)   {Gaussians-to-mask (\ref{subsec:method_mask})\\ $\chi(\mathbf x;\theta)\in[0,1]$}; \\
  \node[lossvis]  (lvis)   {Visual loss\\ $\mathcal L_{\mathrm{vis}}(\theta)=\sum_k \ell(\hat I_k,I_k)$}; &
  \node[physstep] (sim)    {Differentiable\\simulator $\mathcal S$ (\ref{subsec:method_fluid_sim})\\ $\tau(\theta)= \mathcal{S}(\chi) = \{\mathbf u_t,p_t,\rho_t,\dots\}_{t=0}^T$}; \\
  \node[block, draw=none, minimum height=0mm, inner sep=0pt] {}; &
  \node[lossphys] (lphys)  {Physical objective\\ $\mathcal L_{\mathrm{phys}}(\theta)$ $=$ \tcbox[philossbox]{$\Phi(\tau(\theta))$}}; \\
};

% --- Merge + update ---
\node[merge, below=10mm of M] (obj)
{Joint objective (\ref{subsec:method_jointopt})\\[2pt]
$\mathcal J(\theta)=
\tcbox[vislossbox]{$\mathcal L_{\mathrm{vis}}(\theta)$}
+
\tcbox[physlossbox]{$\mathcal L_{\mathrm{phys}}(\theta)$}
+
\mathcal L_{\mathrm{reg}}(\theta)$};

\node[update, below=7mm of obj] (opt)
{Update via Gradient Descent\\ $\theta \leftarrow \theta - \eta\,\nabla_\theta \mathcal J(\theta)$};

% wiring
\draw[visline]  (render) -- (lvis);
\draw[physline] (mask) -- (sim);
\draw[physline] (sim)  -- (lphys);

% background lanes
\begin{pgfonlayer}{background}
  \node[lanevisstyle,  fit=(render)(lvis)(imgin),
        label={[font=\bfseries\small, text=blue!65!black]north:Visual supervision (\ref{subsec:visual_supervision})}] (lanevis) {};
  \node[lanephysstyle, fit=(mask)(sim)(lphys)(desin),
        label={[font=\bfseries\small, text=teal!60!black]north:Physics supervision (\ref{subsec:method_sim})}] (lanephys) {};
\end{pgfonlayer}

% loop back
\coordinate (loopX) at ($(lanevis.west)+(-5mm,0)$);
\coordinate (loopL) at (loopX |- opt.west);
\coordinate (loopU) at (loopL |- theta.west);
\draw[dashline, rounded corners=10pt] (opt.west) -- (loopL) -- (loopX) -- (loopU) -- (theta.west);

% anchor points above lane labels
\path (lanevis.north)  ++(0,6mm) coordinate (lanevisLabelTop);
\path (lanephys.north) ++(0,6mm) coordinate (lanephysLabelTop);

% theta -> lanes
\draw[visline]  (theta.south -| lanevisLabelTop)  -- (lanevisLabelTop);
\draw[physline] (theta.south -| lanephysLabelTop) -- (lanephysLabelTop);

% lanes -> objective
\draw[visline]  (lanevis.south)  -- (lanevis.south  |- obj.north);
\draw[physline] (lanephys.south) -- (lanephys.south |- obj.north);

% gradient flow
\def\gradSep{3mm}
\def\gradLabelX{2mm}

\draw[gradline]
  ($ (obj.north -| lanevis.south) + (-\gradSep,0) $) --
  ($ (lanevis.south)             + (-\gradSep,0) $)
  node[midway, gradlabel, left, xshift=-\gradLabelX]
    {$\frac{\partial \mathcal J}{\partial \mathcal L_{\mathrm{vis}}}$};

\draw[gradline]
  ($ (obj.north -| lanephys.south) + (\gradSep,0) $) --
  ($ (lanephys.south)             + (\gradSep,0) $)
  node[midway, gradlabel, right, xshift=\gradLabelX]
    {$\frac{\partial \mathcal J}{\partial \mathcal L_{\mathrm{phys}}}$};

\draw[gradvis]
  ($ (lvis.north)   + (-\gradSep,0) $) --
  ($ (render.south) + (-\gradSep,0) $)
  node[midway, gradlabel, left, xshift=-\gradLabelX]
    {$\frac{\partial \mathcal L_{\mathrm{vis}}}{\partial \hat I}$};

\draw[gradvis]
  ($ (lanevisLabelTop) + (-\gradSep,0) $) --
  ($ (theta.south -| lanevisLabelTop) + (-\gradSep,0) $)
  node[midway, gradlabel, left, xshift=-\gradLabelX]
    {$\frac{\partial \hat I}{\partial \theta}$};

\draw[gradphys]
  ($ (lphys.north) + (\gradSep,0) $) --
  ($ (sim.south)   + (\gradSep,0) $)
  node[midway, gradlabel, right, xshift=\gradLabelX]
    {$\frac{\partial \mathcal L_{\mathrm{phys}}}{\partial \tau}$};

\draw[gradphys]
  ($ (sim.north)  + (\gradSep,0) $) --
  ($ (mask.south) + (\gradSep,0) $)
  node[midway, gradlabel, right, xshift=\gradLabelX]
    {$\frac{\partial \tau}{\partial F}$};

\draw[gradphys]
  ($ (lanephysLabelTop) + (\gradSep,0) $) --
  ($ (theta.south -| lanephysLabelTop) + (\gradSep,0) $)
  node[midway, gradlabel, right, xshift=\gradLabelX]
    {$\frac{\partial F}{\partial \theta}$};

\end{tikzpicture}

%% file: appendix.tex
\section{Efficient tiled evaluation of the Gaussian-induced solid mask}
\label{app:mask_tiling}

This appendix describes an efficient strategy for evaluating the Gaussian-induced solid
mask defined in Sec.~\ref{subsec:method_mask}.
The method is mathematically equivalent to evaluating the occupancy field
$\chi (\mathbf x;\theta)$ at all grid locations, but exploits the spatial locality of Gaussian support
to reduce computational cost in practice.

\subsection{Motivation}

Evaluating the solid occupancy field
\[
\chi (\mathbf x;\theta) = 1 - \prod_{i=1}^N \big(1-\hat\alpha_i(\mathbf x)\big)
\]
naively at every grid cell $\mathbf x\in\Omega_h$ requires computing contributions from all
$N$ Gaussians at all grid locations, resulting in $O(N|\Omega_h|)$ time complexity.
In typical scenes, however, each Gaussian has compact spatial support relative to the grid,
and most Gaussians do not contribute meaningfully to a given grid region.
We therefore adopt a tiled evaluation strategy inspired by rasterization-style culling in
Gaussian splatting~\cite{kerbl_3d_2023}.

\subsection{Tile-based culling and evaluation}

For each Gaussian $i$, we compute a conservative grid-aligned axis-aligned bounding box (AABB)
that encloses the region where its contribution $\hat\alpha_i(\mathbf x)$ is non-negligible.
In practice, this AABB is obtained from the Gaussian mean $\mu_i$ and covariance $\Sigma_i$
using covariance-derived extents,
\[
\mu_i \pm \kappa\,\sqrt{\mathrm{diag}(\Sigma_i)},
\]
for a fixed constant $\kappa$.

We partition the grid $\Omega_h$ into disjoint cubic tiles of fixed size.
For each tile, we identify the subset of Gaussians whose AABBs intersect the tile.
Only this subset is used when evaluating the occupancy field within the tile.
Within each tile, Gaussian contributions are computed in a batched fashion at all voxel centers
belonging to the tile, and the probabilistic union is applied to obtain the local occupancy values.

\subsection{Algorithm}

Algorithm~\ref{alg:tiled_mask} summarizes the tiled mask evaluation procedure.

\begin{algorithm}[H]
\caption{Tiled evaluation of Gaussian-induced solid mask}
\label{alg:tiled_mask}
\begin{algorithmic}[1]
\Require Gaussian parameters $\{(\mu_i,\Sigma_i,\alpha_i)\}_{i=1}^N$, grid $\Omega_h$, tile size $s$
\Ensure Solid mask $\chi (\mathbf x)\in[0,1]$ for all $\mathbf x\in\Omega_h$
\State Compute conservative AABBs for all Gaussians
\ForAll{tiles $T\subset\Omega_h$}
    \State Identify Gaussians whose AABBs intersect $T$
    \If{no Gaussians intersect $T$}
        \State continue
    \EndIf
    \State Evaluate $\hat\alpha_i(\mathbf x)$ for all $\mathbf x\in T$ and intersecting Gaussians
    \State Compute local solid occupancy
    \[
    \chi_T(\mathbf x)=1-\prod_i\big(1-\hat\alpha_i(\mathbf x)\big)
    \]
\EndFor
\end{algorithmic}
\end{algorithm}

\subsection{Differentiability and complexity}

All operations within each tile—Gaussian evaluation, aggregation via the product form, and assignment
to the mask—are differentiable with respect to the Gaussian parameters.
The tiled structure introduces no discontinuities, as each grid cell is evaluated independently
using a fixed set of contributing Gaussians determined by conservative AABB culling.

Let $\bar P$ denote the average number of Gaussians overlapping a tile.
The expected runtime of the tiled evaluation scales as
$O(|\Omega_h|\bar P)$, with $\bar P\ll N$ in practice.
This substantially reduces computation compared to naive dense evaluation while leaving the
mathematical definition of the occupancy and solid mask unchanged.

\subsection{Additional Teapot Results}
\label{app:teapot_results}

\begin{figure}[p] % 'p' forces it to a dedicated page
\centering
\begin{tabular}{@{} c c c c c @{}}
\scriptsize \textbf{\shortstack{Ground\\Truth}} &
  \scriptsize \textbf{\shortstack{SDF\\Render}} &
  \scriptsize \textbf{\shortstack{NeRF\\Render}} &
  \scriptsize \textbf{\shortstack{3DGS\\Render}} &
  \scriptsize \textbf{\shortstack{\ProjectShortName\\(Ours) Render}} \\[1pt]
  
  \RenderRow{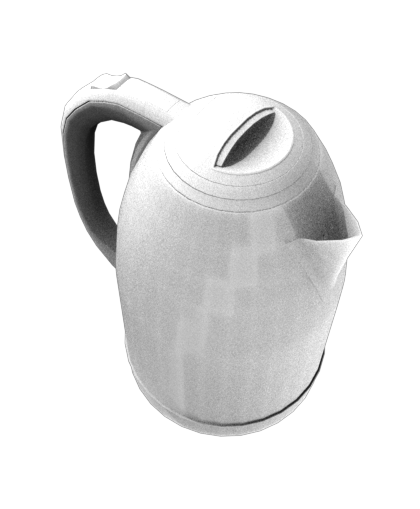}
  % \RenderRow{01}
  % \RenderRow{02}
  % \RenderRow{03}
  % \RenderRow{04}
  % Split here if the page is too full
  % \RenderRow{05}
  % \RenderRow{06}
  \RenderRow{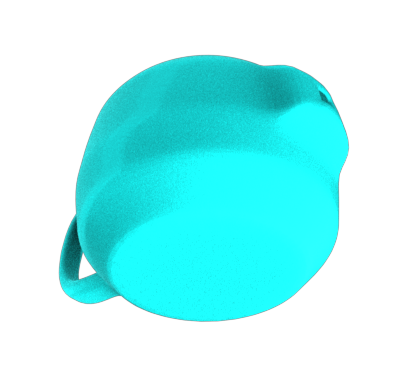}
  \RenderRow{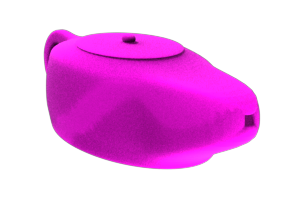}
  \RenderRow{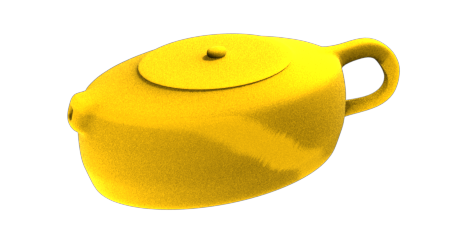}
  \RenderRow{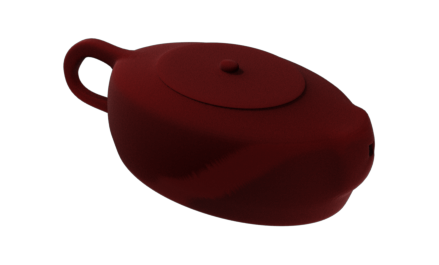}
  \RenderRow{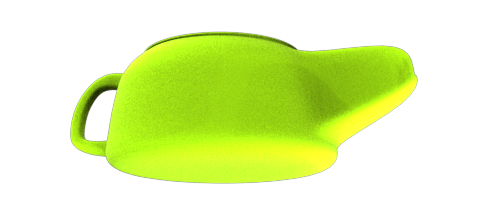}
  % \RenderRow{12}
  % \RenderRow{13}
  % \RenderRow{14}

\end{tabular}
\caption{Full Render Comparison (Teapots 07--12). All methods are rendered from the test viewpoint. We can see from this image that our method produces visually comparable results to those of the baseline, indicating that our physics objective does not substantially degrade visual quality.}
\label{fig:render_results_appendix_1}
\end{figure}

\begin{figure}[p] % 'p' forces it to a dedicated page
\centering
\begin{tabular}{@{} c c c c c @{}}
\scriptsize \textbf{\shortstack{Ground\\Truth}} &
  \scriptsize \textbf{\shortstack{SDF\\Render}} &
  \scriptsize \textbf{\shortstack{NeRF\\Render}} &
  \scriptsize \textbf{\shortstack{3DGS\\Render}} &
  \scriptsize \textbf{\shortstack{\ProjectShortName\\(Ours) Render}} \\[1pt]
  
  % \RenderRow{00}
  % \RenderRow{01}
  % \RenderRow{02}
  % \RenderRow{03}
  % \RenderRow{04}
  % Split here if the page is too full
  % \RenderRow{05}
  % \RenderRow{06}
  % \RenderRow{07}
  % \RenderRow{08}
  % \RenderRow{09}
  % \RenderRow{10}
  % \RenderRow{11}
  \RenderRow{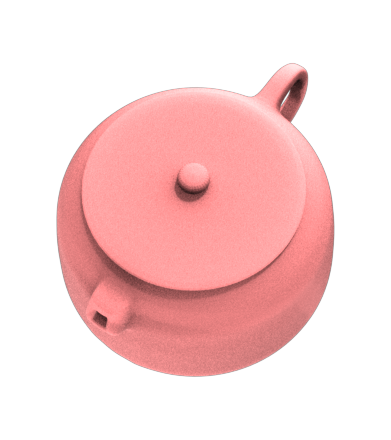}
  \RenderRow{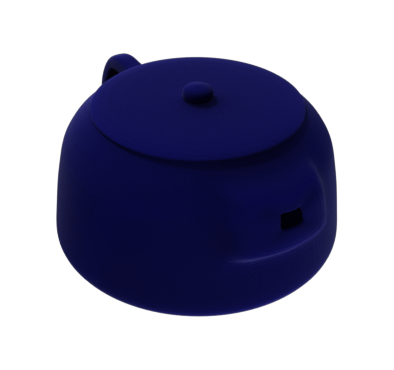}
  \RenderRow{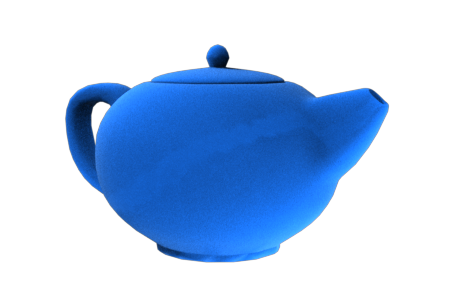}

\end{tabular}
\caption{Full Render Comparison (Teapots 13--15). All methods are rendered from the test viewpoint. We can see from this image that our method produces visually comparable results to those of the baseline, indicating that our physics objective does not substantially degrade visual quality.}
\label{fig:render_results_appendix_2}
\end{figure}

\begin{figure}[p]
\centering
\begin{tabular}{@{} c c c c c @{}}
\scriptsize \textbf{\shortstack{Reference\\View}} &
\scriptsize \textbf{\shortstack{SDF\\Cutaway}} &
\scriptsize \textbf{\shortstack{NeRF\\Cutaway}} &
\scriptsize \textbf{\shortstack{3DGS\\Cutaway}} &
\scriptsize \textbf{\shortstack{\ProjectShortName\\(Ours) Cutaway}} \\[1pt]

  \CutawayRow{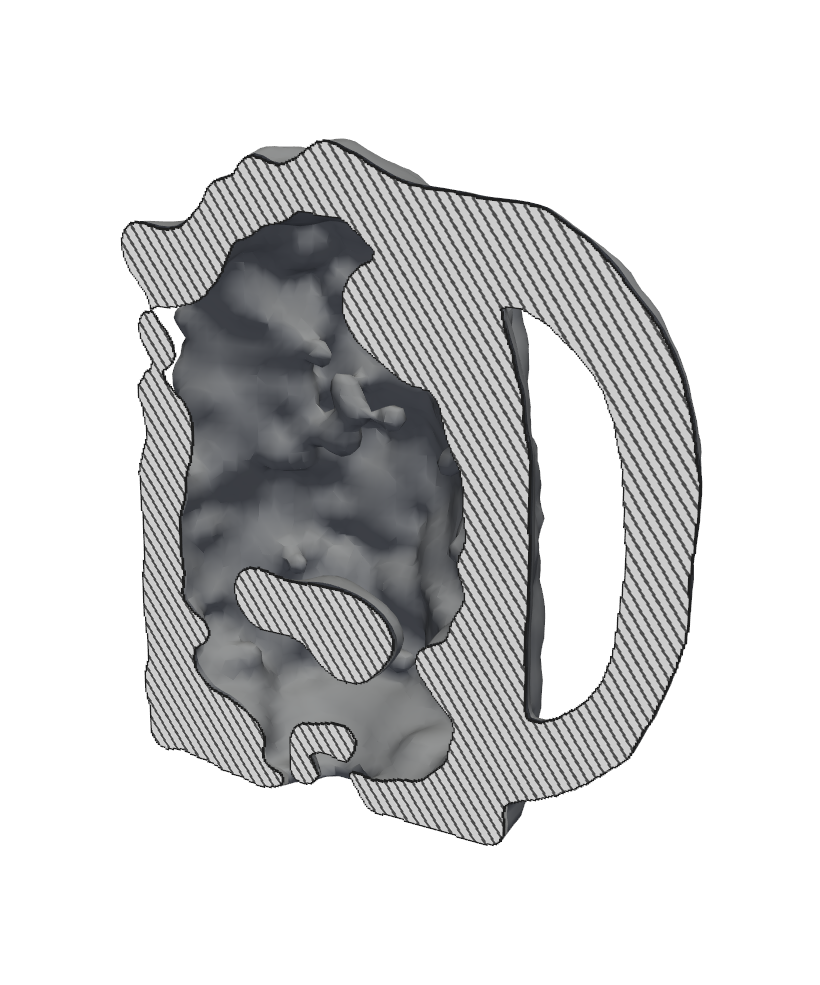}
  % \SliceRow{01}
  % \SliceRow{02}
  % \SliceRow{03}
  % \SliceRow{04}
  % \SliceRow{05}
  % \SliceRow{06}
  \CutawayRow{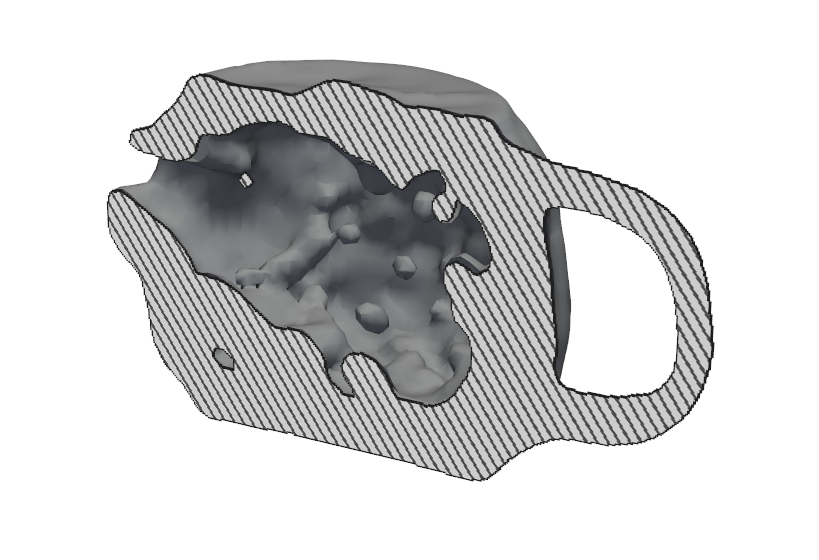}
  \CutawayRow{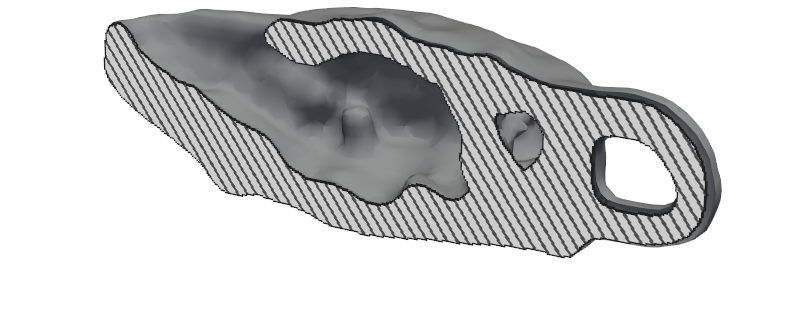}
  \CutawayRow{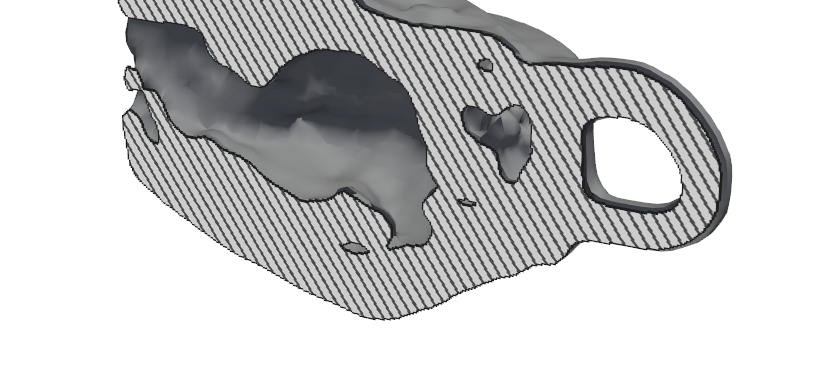}
  \CutawayRow{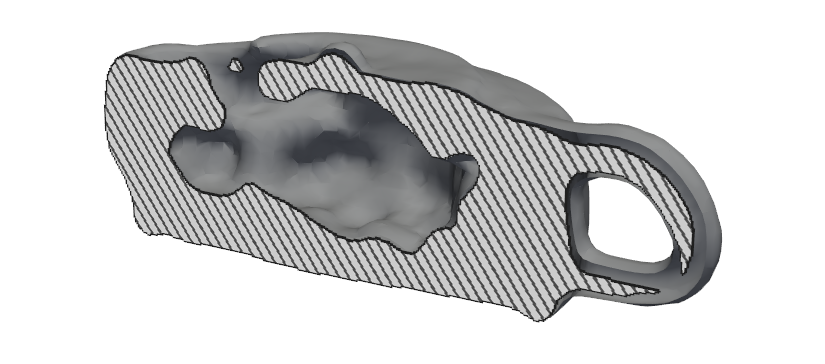}
  \CutawayRow{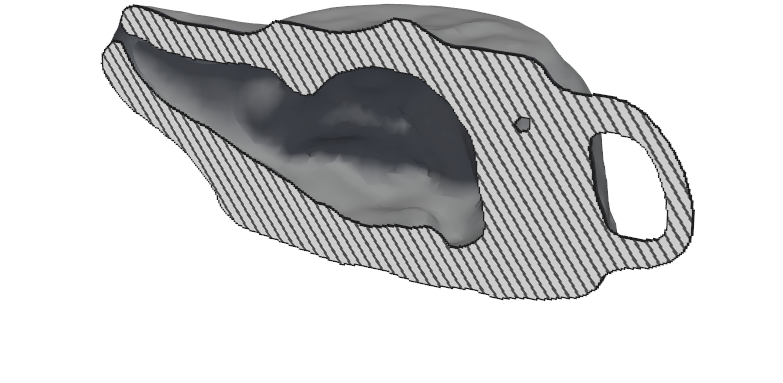}
  % \SliceRow{12}
  % \SliceRow{13}
  % \SliceRow{14}

\end{tabular}
\caption{Geometry cutaway Comparison (Teapots 07--12). The first column is the reference image; subsequent columns show the internal geometry extracted from the density field. Note that our method produces a clear path from an internal cavity to the outside of the object.}
\label{fig:cross_section_results_appendix_1}
\end{figure}

\begin{figure}[p]
\centering
\begin{tabular}{@{} c c c c c @{}}
\scriptsize \textbf{\shortstack{Reference\\View}} &
\scriptsize \textbf{\shortstack{SDF\\Cutaway}} &
\scriptsize \textbf{\shortstack{NeRF\\Cutaway}} &
\scriptsize \textbf{\shortstack{3DGS\\Cutaway}} &
\scriptsize \textbf{\shortstack{\ProjectShortName\\(Ours) Cutaway}} \\[1pt]
  
  % \SliceRow{00}
  % \SliceRow{01}
  % \SliceRow{02}
  % \SliceRow{03}
  % \SliceRow{04}
  % \SliceRow{05}
  % \SliceRow{06}
  % \SliceRow{07}
  % \SliceRow{08}
  % \SliceRow{09}
  % \SliceRow{10}
  % \SliceRow{11}
  \CutawayRow{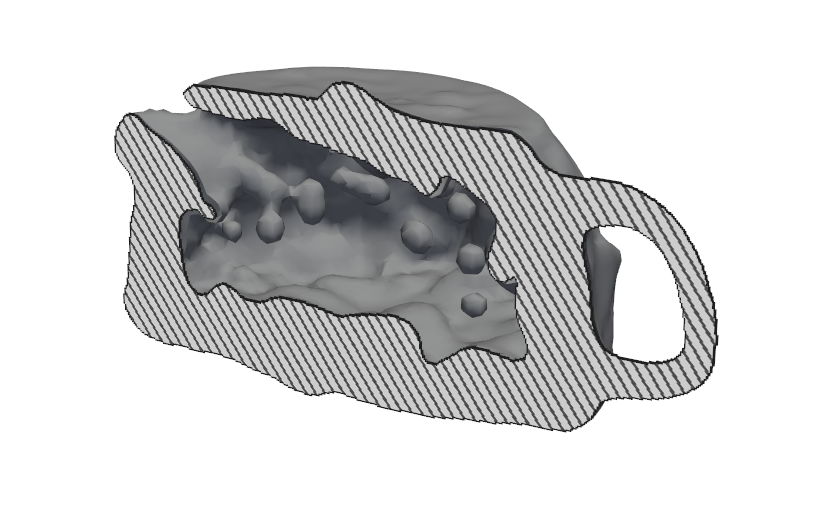}
  \CutawayRow{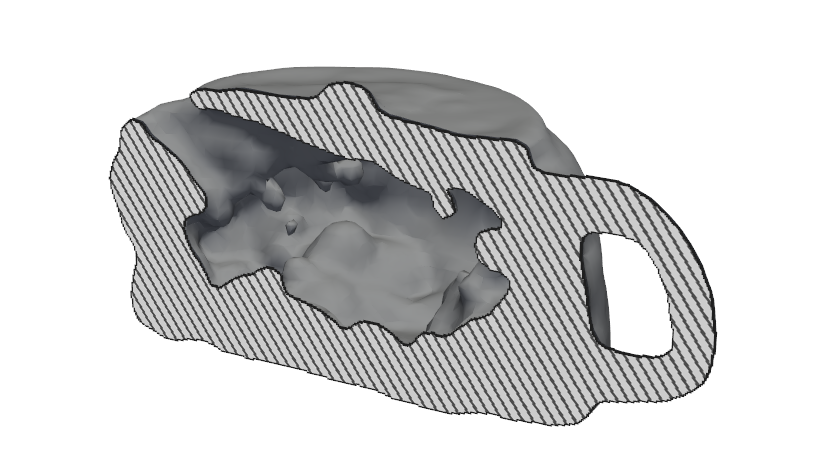}
  \CutawayRow{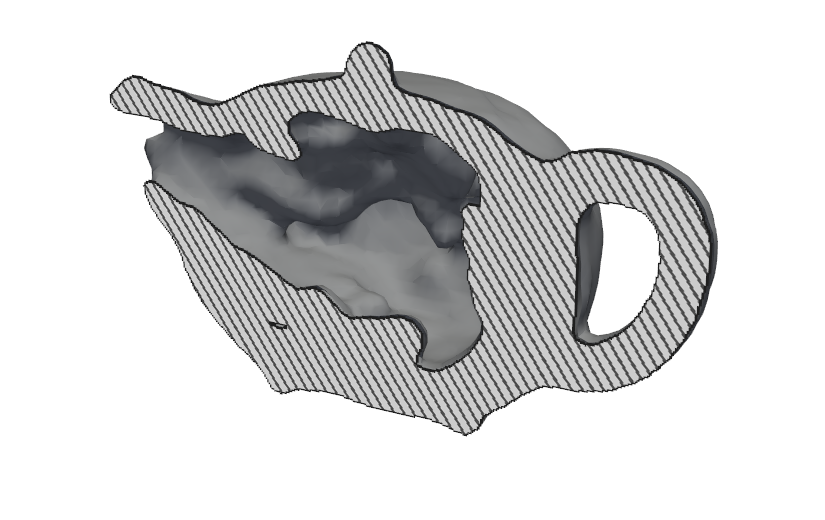}

\end{tabular}
\caption{Geometry cutaway Comparison (Teapots 13--15). The first column is the reference image; subsequent columns show the internal geometry extracted from the density field. Note that our method produces a clear path from an internal cavity to the outside of the object.}
\label{fig:cross_section_results_appendix_2}
\end{figure}

\begin{figure}[p]
\centering
\begin{tabular}{@{} c c c c c @{}}
\scriptsize \textbf{\shortstack{Reference\\View}} &
\scriptsize \textbf{\shortstack{SDF\\Simulation}} &
\scriptsize \textbf{\shortstack{NeRF\\Simulation}} &
\scriptsize \textbf{\shortstack{3DGS\\Simulation}} &
\scriptsize \textbf{\shortstack{\ProjectShortName (Ours)\\Simulation}} \\[1pt]
  
  \SimRow{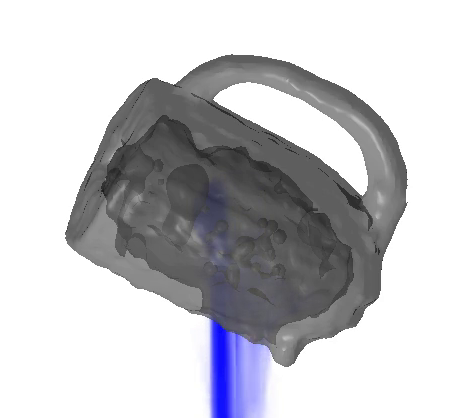}
  % \SimRow{01}
  % \SimRow{02}
  % \SimRow{03}
  % \SimRow{04}
  % \SimRow{05}
  % \SimRow{06}
  \SimRow{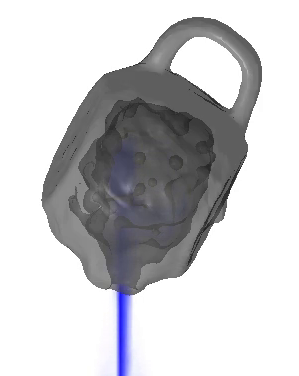}
  \SimRow{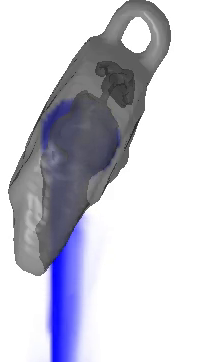}
  \SimRow{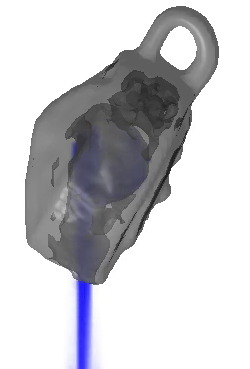}
  \SimRow{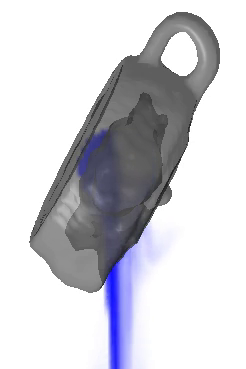}
  \SimRow{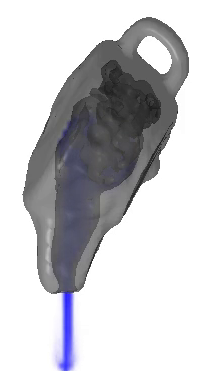}
  % \SimRow{12}
  % \SimRow{13}
  % \SimRow{14}

\end{tabular}
\caption{Physics Simulation Comparison (Teapots 07--12). Snapshot taken at final frame of each simulation. The first column is the static reference image. We can see that our method allows fluid (blue) to be poured out from the teapot, while baselines tend to trap it inside.}
\label{fig:simulation_results_appendix_1}
\end{figure}

\begin{figure}[p]
\centering
\begin{tabular}{@{} c c c c c @{}}
\scriptsize \textbf{\shortstack{Reference\\View}} &
\scriptsize \textbf{\shortstack{SDF\\Simulation}} &
\scriptsize \textbf{\shortstack{NeRF\\Simulation}} &
\scriptsize \textbf{\shortstack{3DGS\\Simulation}} &
\scriptsize \textbf{\shortstack{\ProjectShortName (Ours)\\Simulation}} \\[1pt]
  
  % \SimRow{00}
  % \SimRow{01}
  % \SimRow{02}
  % \SimRow{03}
  % \SimRow{04}
  % \SimRow{05}
  % \SimRow{06}
  % \SimRow{07}
  % \SimRow{08}
  % \SimRow{09}
  % \SimRow{10}
  % \SimRow{11}
  \SimRow{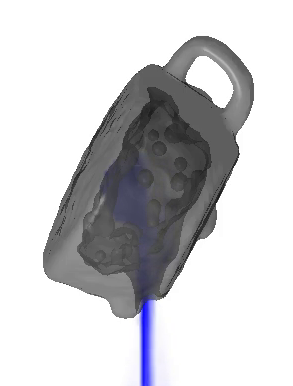}
  \SimRow{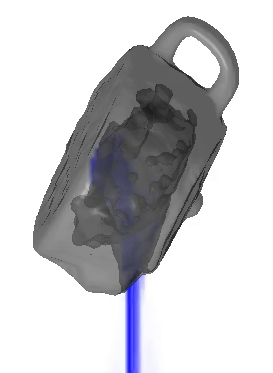}
  \SimRow{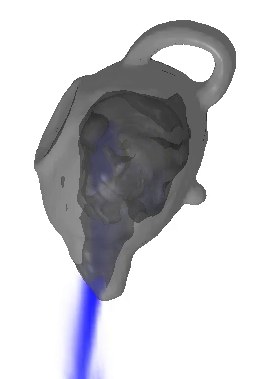}

\end{tabular}
\caption{Physics Simulation Comparison (Teapots 13--15). Snapshot taken at final frame of each simulation. The first column is the static reference image. We can see that our method allows fluid (blue) to be poured out from the teapot, while baselines trap it inside.}
\label{fig:simulation_results_appendix_2}
\end{figure}